\documentclass[11pt]{article}
\pdfoutput=1
\usepackage[utf8]{inputenc} 
\usepackage[T1]{fontenc} 
\usepackage{times}
\usepackage{url}
  \newlength\titlebox 
\usepackage{pdfpages}

\begin{document}

  \begin{center}
  \LARGE WeaSuL 2021\\
  \vspace*{55mm}
    {\bf
    \fontsize{14}{14}\selectfont
    \vspace{2cm}
    \LARGE
    Proceedings of the First Workshop on Weakly Supervised Learning (WeaSuL) \\
    \vspace{2cm}
    \hspace*{1cm}} \\ 
    \vspace{43mm}
    \LARGE
    May 7, 2021 \\
    Co-located with ICLR (Online)
  \end{center}
  
  \newpage

\begin{center}
  {\Large \bf Introduction}
\end{center}

\vspace*{0.5cm}

Welcome to WeaSuL 2021, the First Workshop on Weakly Supervised Learning, co-located with ICLR 2021. 

Deep learning relies on massive training sets of labeled examples to learn from - often tens of thousands to millions to reach peak predictive performance, but large amounts of training data are only available for very few standardized learning problems. Even small variations of the problem specification or changes in the data distribution would necessitate re-annotation of large amounts of data.

However, domain knowledge can often be expressed by sets of prototypical descriptions: For example, vision experts can exploit meta information for image labeling, linguists can describe discourse phenomena by prototypical realization patterns, social scientists can specify events of interest by characteristic key phrases, and bio-medical researchers have databases of known interactions between drugs or proteins that can be used for heuristic labeling. These knowledge-based descriptions can be either used as rule-based predictors or as labeling functions for providing partial data annotations. The growing field of weak supervision provides methods for refining and generalizing such heuristic-based annotations in interaction with deep neural networks and large amounts of unannotated data.

In this workshop, we want to advance theory, methods and tools for allowing experts to express prior coded knowledge for automatic data annotations that can be used to train arbitrary deep neural networks for prediction. The ICLR 2021 Workshop on Weak Supervision aims at advancing methods that help modern machine-learning methods to generalize from knowledge provided by experts, in interaction with observable (unlabeled) data.

We called for both long and short papers and received 26 submissions, all of which were double-blindly reviewed by a pool of 29 reviewers. In total, 15 papers were accepted. All the accepted contributions are listed in these Proceedings and those submitted as archival are included in full text.

Learning with weak supervision is both studied from a theoretical perspective as well as applied to a variety of tasks from areas like natural language processing and computer vision. Therefore, the workshop brought together researchers from a wide range of fields, also bridging innovations from academia and the requirements of industry settings.

The program of the workshop, besides 3 oral paper presentations and 12 posters in 2 poster sessions, included invited talks by Marine Carpuat, Heng Ji, Lu Jiang, Dan Roth and Paroma Varma. It closed with a panel discussion with the invited speakers. Snorkel AI provided funding to sponsor ICLR registrations to increase diversity. 

The WeaSuL Workshop Organizers

\newpage


\begin{description}
\item{\bf Organizers:}\vspace{2mm} \\
Michael A. Hedderich, Saarland University (Germany) \\
Benjamin Roth, University of Vienna (Austria)\\
Katharina Kann, University of Colorado Boulder (USA) \\
Barbara Plank, IT University of Copenhagen (Denmark)\\
Alex Ratner, University of Washington (USA)\\
Dietrich Klakow, Saarland University (Germany)\\

\item{\bf Program Committee:}\vspace{2mm} \\
Abhijeet Awasthi, Indian Institute of Technology Bombay\\ 
Andreas Baumann, University of Vienna\\ 
Bo Han, Hong Kong Baptist University\\ 
Chaojun Xiao, Tsinghua University\\ 
Curtis G. Northcutt, Massachusetts Institute of Technology\\ 
Daniel Y. Fu, Stanford University\\ 
David Adelani, Saarland University\\ 
Dawei Zhu, Saarland University\\ 
Edwin Simpson, University of Bristol\\ 
Erion Çano, University of Vienna\\ 
Ivan Habernal, TU Darmstadt\\ 
Jacob Goldberger, Bar-Ilan Univeristy\\ 
Judith Gaspers, Amazon\\ 
Julia Hockenmaier, University of Illinois at Urbana-Champaign\\ 
Khaled K. Saab, Stanford University\\ 
Lukas Lange, Bosch Center for Aritificial Intelligence\\ 
Mahaveer Jain, Facebook\\ 
Marina Speranskaya, LMU Munich\\ 
Nils Rethmeier, DFKI Berlin\\ 
Pierre Lison, University of Oslo\\ 
Quanming Yao, 4Paradigm\\ 
Sarah Hooper, Stanford University\\ 
Seffen Eger, TU Darmstadt\\ 
Shiran Dudy, Oregon Health \& Science University\\ 
Stephen H. Bach, Brown University\\ 
Thomas Trost, Saarland University\\ 
Tongliang Liu, University of Sydney\\ 
Vincent Chen, Snorkel AI\\ 
Xiang Dai, University of Sydney\\

\vspace{3mm}
\item{\bf Invited Speaker:}\vspace{2mm} \\
Marine Carpuat, University of Maryland \\
Heng Ji, University of Illinois at Urbana-Champaign \\
Lu Jiang, Google Research \\
Dan Roth, University of Pennsylvania \\
Paroma Varma, Snorkel AI 

\end{description}

\newpage

\setlength{\parindent}{0in}
\setlength{\parskip}{2ex}
\renewcommand{\baselinestretch}{0.87}

\begin{center}
{\Large \bf
  Accepted Papers
}
\end{center}
\textit{TADPOLE: Task ADapted Pre-training via anOmaLy dEtection}\\
Vivek Madan, Ashish Khetan and Zohar Karnin\\ \\ 
\textit{CIGMO: Learning categorical invariant deep generative models from grouped data}\\
Haruo Hosoya\\ \\ 
\textit{Handling Long-Tail Queries with Slice-Aware Conversational Systems}\\
Cheng Wang, Sun Kim, Taiwoo Park,Sajal Choudhary, Sunghyun Park, Young-Bum Kim, Ruhi Sarikaya and Sungjin Lee\\ \\ 
\textit{Tabular Data Modeling via Contextual Embeddings}\\
Xin Huang, Ashish Khetan, Milan Cvitkovic and Zohar Karnin\\
\url{https://arxiv.org/abs/2012.06678} \\ \\ 
\textit{Pre-Training by Completing Points Cloud}\\
Hanchen Wang, Liu Qi, Xiangyu Yue, Matt Kusner and Joan Lasenby | (non-archival)\\ \\ 
\textit{AutoTriggER: Named Entity Recognition with Auxiliary Trigger Extraction}\\
Dong-Ho Lee, Ravi Kiran Selvam, Sheikh Muhammad Sarwar, Bill Yuchen Lin, Fred Morstatter, Jay Pujara, Elizabeth Boschee, James Allan and Xiang Ren | (non-archival)\\ \\ 
\textit{Active WeaSuL: Improving Weak Supervision with Active Learning}\\
Samantha R Biegel, Rafah El-Khatib, Luiz Otavio Vilas Boas Oliveira, Max Baak and Nanne Aben\\
\url{https://arxiv.org/abs/2104.14847} \\ \\ 
\textit{Dependency Structure Misspecification in Multi-Source Weak Supervision Models}\\
Salva Rühling Cachay, Benedikt Boecking and Artur Dubrawski\\ 
\url{https://arxiv.org/abs/2106.10302} \\ \\ 
\textit{Weakly-Supervised Group Disentanglement using Total Correlation}\\
Linh Tran, Saeid Asgari Taghanaki, Amir Hosein Khasahmadi, Aditya Sanghi\\ \\ 
\textit{Better adaptation to distribution shifts with Robust Pseudo-Labeling}\\
Evgenia Rusak, Steffen Schneider, Peter Gehler, Oliver Bringmann, Bernhard Schölkopf, Wieland Brendel and Matthias Bethge | (non-archival)\\ \\ 
\textit{Transformer Language Models as Universal Computation Engines}\\
Kevin Lu, Aditya Grover, Pieter Abbeel and Igor Mordatch | (non-archival)\\ \\ 
\textit{Using system context information to complement weakly labeled data}\\
Matthias Meyer, Michaela Wenner, Clément Hibert, Fabian Walter and Lothar Thiele\\ \\ 
\textit{Is Disentanglement all you need? Comparing Concept-based \& Disentanglement Approaches}\\
Dmitry Kazhdan, Botty Dimanov, Helena Andres Terre, Pietro Lió, Mateja Jamnik and Adrian Weller | (non-archival)\\ \\ 
\textit{Weakly Supervised Multi-task Learning for Concept-based Explainability}\\
Vladimir Balayan, Catarina G Belém, Pedro Saleiro and Pedro Bizarro\\
\url{https://arxiv.org/abs/2104.12459} \\ \\ 
\textit{Pervasive Label Errors in Test Sets Destabilize Machine Learning Benchmarks}\\
Curtis G Northcutt, Anish Athalye and Jonas Mueller \\
\url{https://arxiv.org/abs/2103.14749} 

\newpage



\section*{Supplementary material}

# TADPOLE: TASK ADAPTED PRE-TRAINING VIA ANOMALY DETECTION

**Vivek Madan**
AWS AI Labs
vivmadan@amazon.com

**Ashish Khetan**
AWS AI Labs
khetan@amazon.com

**Zohar Karnin**
AWS AI Labs
zkarnin@amazon.com

## ABSTRACT

The paradigm of pre-training followed by finetuning has become a standard procedure for NLP tasks, with a known problem of domain shift between the pre-training and downstream corpus. Previous works have tried to mitigate this problem with additional pre-training, either on the downstream corpus itself when it is large enough, or on a manually curated unlabeled corpus of a similar domain. In this paper, we address the problem for the case when the downstream corpus is too small for additional pre-training. We propose TADPOLE, a task adapted pre-training framework based on data selection techniques adapted from *Domain Adaptation*. We formulate the data selection as an anomaly detection problem that unlike existing methods works well when the downstream corpus is limited in size. It results in a scalable and efficient unsupervised technique that eliminates the need for any manual data curation. We evaluate our framework on eight tasks across four different domains: Biomedical, Computer Science, News, and Movie reviews, and compare its performance against competitive baseline techniques from the area of Domain Adaptation. Our framework outperforms all the baseline methods. On large datasets we get an average gain of $0.3\%$ in performance but on small datasets with less than 5K training examples, we get a much higher gain of $1.8\%$. This shows the efficacy of domain adapted finetuning when the task dataset is small.

## 1 INTRODUCTION

Pre-trained language models such as ELMo (Peters et al., 2018), GPT (Radford et al., 2018), BERT (Devlin et al., 2018), Transformer-xl (Dai et al., 2019) and XLNet (Yang et al., 2019) have become a key component in solving virtually all natural language tasks. These models are pre-trained on large amount of cross-domain data ranging from Wikipedia to Book corpus to news articles to learn powerful representations. A generic approach for using these models consists of two steps: (a) Pre-training: train the model on an extremely large general domain corpus, e.g. with masked language model loss; (b) Finetuning: finetune the model on labeled task dataset for the downstream task.

Even though the approach of pre-training followed by fine-tuning has been very successful, it suffers from *domain shift* when applied to tasks containing text from a domain that is not sufficiently represented in the pre-training corpus. An immediate way of solving the problem is to pre-train the model on task domain data instead of the general domain data. For a handful of very popular task domains, the research community invested time and resources to collect a large domain-specific data corpus and pre-train a language model on it. The models include BioBERT pre-trained on biomedical text (Lee et al., 2020), ClinicalBERT pre-trained on clinical notes (Huang et al., 2019), SciBERT pre-trained on semantic scholar corpus (Beltagy et al., 2019), and FinBERT pre-trained on financial documents (Araci, 2019). These models achieve significant gain in performance over a model trained on general domain data, when the downstream task belongs to the respective domains.

These papers demonstrate how useful it can be to shift the domain of the pre-trained model. However, the approach is expensive and time consuming as it requires collecting gigabytes of domain data for each new task. The long-tail of domains remains left behind without a realistic solution. To mitigate this, in absence of the huge task domain data, a different known approach is to collect a medium (MBs, not GBs) amount of unlabeled task data, and adapt the pre-trained (on general data) model by e.g. extending the pre-training procedure on the unlabeled data (Howard & Ruder, 2018; Gururangan

et al., 2020). Such task adapted pre-training approach achieves relatively smaller gain but is less expensive. Although this approach is cheaper in terms of manual labor when compared to domain adapted BERT, it still requires an effort to collect unlabeled data. It requires much more data than what is needed for only fine-tuning. This is often impossible to achieve, for example when data is highly sensitive. In this paper we propose a solution to this challenging problem, providing domain adaptation to the pre-trained model without the need for any manual effort of data collection.

The high level idea is quite intuitive. Given a generic pre-training data containing text from multiple domains, we filter the available general domain to contain only pieces that are similar to the downstream task corpus. By continuing the pre-training process on this adapted corpus we achieve a better tuned pre-trained model. Figure 1 illustrate the feasibility of this approach with an example downstream task from a medical domain and highlighted text from a news article available in a general domain corpus. The key for a successful implementation is finding the best way of evaluating the similarity of a given snippet to the downstream task.

| RCT20K TASK DATA | News Article |
|---|---|
| To investigate the efficacy of 6 weeks of daily low-dose oral prednisolone in improving pain, mobility, and systemic los-grade . . . [OBJECTIVE] <br> A total of 125 patients with primary knee OA were randomize . . . [METHODS] <br> Outcome measures included pain reduction and systemic inflammation markers . . . [METHODS] | For as much as we workout warriors recite that whole "no pain, no gain" mantra, we sure do pop a lot of painkillers. A recent article published in. . . These popular medicines, known as non-steroidal anti-inflammatory drugs, or NSAIDs, work by suppressing inflammation. . . the article kind of blows past is the fact plenty of racers . . . |

Figure 1: Identification of task-data (top panel, medical data) in general domain corpus (bottom panel).

Although not many methods exist to solve the problem of domain shift in the context of pretraining, literature on Domain Adaptation provides several methods for the core task of evaluating the above mentioned similarity. These previous approaches use either a simple language model (LM) (Moore & Lewis, 2010; Axelrod et al., 2011; Duh et al., 2013; Wang et al., 2017b; van der Wees et al., 2017), or a hand crafted similarity score (Wang et al., 2017a; Plank & Van Noord, 2011; Remus, 2012; Van Asch & Daelemans, 2010). The LM based technique are often both over simplistic, and require a fairly large corpus of task data to create a reasonable LM. The hand crafted similarity scores can be seen as ad-hoc methods for distinguishing inliers from outliers (i.e., anomaly detection); they tend to be focused on individual tasks and do not generalize well.

We formulate the similarity evaluation task as that of anomaly detection and propose a Task ADapted Pre-training via anOmaLy dEtection (TADPOLE) framework. Indeed, anomaly detection methods given a domain of instance are able to provide a score for new instances assessing how likely they are to belong to the input domain. We exploit pre-trained models to get sentence representations that are in turn used to train an anomaly detection model. By using pre-trained models, our method is effective even for small text corpora. By taking advantage of existing anomaly detection methods, we replace hand-crafted rules with techniques proven to generalize well.

In what follows we discuss how we implement our technique and compare it with other data selection methods based on extensive experimental results. We start by filtering out the subset of general domain corpus most relevant to the task. To do this, we explore several anomaly detection methods and give a quantitative criterion to identify the best method for a given task data. Then, we start with a pre-trained model on the general domain corpus and run additional pre-training for only 5% more steps on the filtered corpus from different methods. This is followed by the regular finetuning on the labeled task data. We measure the performance gain as an improvement in accuracy of finetuned model with additional pre-training vs the accuracy of finetuned model without additional pre-training. To establish the performance gain of TADPOLE, we evaluate it on eight tasks across four domains: Biomedical, Computer Science, News, and Movie reviews. We investigate all aspects of TADPOLE by comparing its performance with various baselines based on its variants and the competitive methods available in literature. The main highlights of our work are as follows:

- We provide TADPOLE, a novel anomaly detection based framework for adapting pre-training for the downstream task. The framework is explained in detail and all its steps are justified via extensive ablation studies.

- TADPOLE is superior to all the baseline methods including (i) LM based relevance score (iii) Distance based relevance score (iii) Continued pre-training on the task data.
- Our method achieves an average $1.07\%$ lift in accuracy over eight tasks from four different domains whereas the baselines achieve no more than $0.34\%$ gain. In all individual tasks, our method is either on par or (statistical) significantly better than all alternatives.
- For tasks with small labeled dataset (less than 5K examples), our method achieves an even higher average lift of $1.82\%$ in accuracy.
- For a task requiring little domain adaptation, GLUE sentiment analysis, our method achieves an improvement of $0.4\%$ in accuracy.

## 2 RELATED WORK

Since our focus is on Data Selection Methods, we only discuss the related work on Data Selection in Domain Adaptation here. We discuss the other Domain Adaptation techniques in Appendix A.

**Data Selection:** As discussed above, core of data selection is to determine the relevance weights that in turn modify the source domain to become more similar to the target domain. There has been a sequence of works in trying to find the relevance weights via language models Moore & Lewis (2010); Wang et al. (2017b); van der Wees et al. (2017). For instance, Moore & Lewis (2010), Axelrod et al. (2011) and Duh et al. (2013) train two language models, an in-domain language model on the target domain dataset (same as task domain in our case) and an out-of-domain language model on (a subset of) general domain corpus. Then, relevance score is defined as the difference in the cross-entropy w.r.t. two language models. These methods achieve some gain but have a major drawback. A crucial assumption they rely on: *there is enough in-domain data to train a reasonable in-domain language model.* This assumption is not true in most cases. For most tasks, we only have access to a few thousands or in some cases a few hundreds of examples which is not enough to train a reasonably accurate language model. Our techniques rely on text representations based on the available pre-trained model. As such, our similarity score does not rely on models that can be trained with a small amount of data.

Another line of work defines hand crafted domain similarity measures to assign relevance score and filter out text from a general domain corpus (Wang et al., 2017a; Plank & Van Noord, 2011; Remus, 2012; Van Asch & Daelemans, 2010; Gururangan et al., 2020). For instance, Wang et al. (2017a) define the domain similarity of a sentence as the difference between Euclidean distance of the sentence embedding from the mean of in-domain sentence embeddings and the mean of out-of-domain sentence embeddings. Plank & Van Noord (2011) and Remus (2012) define the similarity measure as the Kullback-Leibler (KL) divergence between the relative frequencies of words, character tetra-grams, and topic models. Van Asch & Daelemans (2010) define domain similarity as Rényi divergence between the relevant token frequencies. These are adhoc measures suitable only for the respective tasks, and can be seen as a manual task-optimized anomaly detection. They fail to generalize well for new tasks and domains. Ruder & Plank (2017) attempts to remedy this issue and tries to learn the correct combination of these metrics for each task. They learn the combination weight vector via Bayesian optimization. However, Bayesian optimization is infeasible for deep networks like BERT. Each optimization step of this process amounts to pre-training the model and finetuning it for the task. For Bayesian optimization to work well it requires repeating this process multiple times, which is prohibitively computationally expensive. Thus, they use models such as linear SVM classifier and LDA which do not yield state-of-the-art performance. In contrast, we propose a lightweight method - based on anomaly detection - that can be applied to state-of-the-art deep language models like BERT.

## 3 TADPOLE: TASK ADAPTED PRE-TRAINING VIA ANOMALY DETECTION

**Language model and Downstream Tasks.** A generic approach for using state-of-the-art language models such as ELMo, GPT, BERT, and XLNet is to pre-train them on an extremely large general domain corpus and then finetune the pre-trained model on the downstream labeled task data. There is evident correlation between model's pre-training loss and its performance on the downstream task after finetuning (Devlin et al., 2018). Our design is motivated by an observation, backed by empirical evidence, that the correlation is even stronger if we consider the pre-training loss not on the pre-training data but the downstream task data.

To make this distinction formal, let $\mathcal{D}, \mathcal{D}_{\text{in}}$ be the pre-training and task data. Let $\Theta$ denote the parameters of the language model and $\ell_{\text{LM}}$ denote the language model loss function. The pre-training loss on pre-training data ($L_{\text{LM}}(\Theta)$) and target data ($L_{\text{LM}}^{\text{in}}(\Theta)$) are defined as follows: $L_{\text{LM}}(\Theta) = \sum_{x \in \mathcal{D}} \ell_{\text{LM}}(x; \Theta), L_{\text{LM}}^{\text{in}}(\Theta) = \sum_{x \in \mathcal{D}_{\text{in}}} \ell_{\text{LM}}(x; \Theta)$. To show that $L_{\text{LM}}^{\text{in}}(\Theta)$ is better correlated with the performance of the downstream task we consider several BERT language models pre-trained on random combinations of datasets from different domains mentioned in Section 4. These models are selected such that *for all the models $L_{LM}(\Theta)$ is roughly the same.* For each model $\Theta$, we estimate $L_{\text{LM}}^{\text{in}}(\Theta)$ and contrast it with the accuracy/f1 score of the finetuned model on the task data.

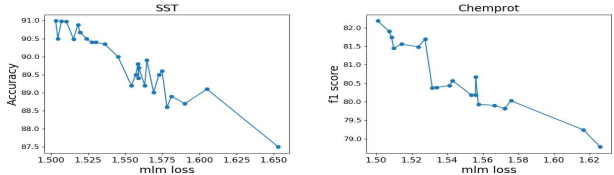

Figure 2: MLM loss of pre-trained BERT on the *task data* vs f1 score of corresponding finetuned BERT. Different points correspond to different BERT models, pre-trained on random combination of different datasets. MLM loss on the general domain corpus for all the pre-trained BERT models considered here is roughly the same.

Figure 2 provides the plots corresponding to this experiment and shows clear evidence that if the language model performs better on the task domain data, then the performance (accuracy/f1 score) of the finetuned model improves. We conclude that in order to ensure success in the downstream task, we should aim to minimize $L_{\text{LM}}^{\text{in}}(\Theta)$. A first attempt would be to pre-train or finetune the language model on $\mathcal{D}_{\text{in}}$. However, training a language model such as ELMo, GPT, BERT or XLNet requires a large corpus with several GBs of text and the available domain specific corpus $\mathcal{D}_{\text{in}}$ is often just the task data which has few MBs of text. Training on such a small dataset would introduce high variance. We reduce this variance by taking training examples from the general domain corpus $\mathcal{D}$, but control the bias this incurs by considering only elements having high relevance to the domain $\mathcal{D}_{\text{in}}$. Formally, we optimize a weighted pre-training loss function

$$L_{\text{LM}}^{\Lambda}(\Theta) = \sum_{x \in D} \lambda(x, \mathcal{D}_{\text{in}}) \cdot \ell_{\text{LM}}(x; \Theta), \quad (1)$$

where $\lambda(x, \mathcal{D}_{\text{in}})$ are relevance weights of instance $x$ for domain $\mathcal{D}_{\text{in}}$. $\lambda(x, \mathcal{D}_{\text{in}})$ is (close to) 1 if $x$ is relevant to $\mathcal{D}_{\text{in}}$ and (close to) 0 otherwise. We compute these weights using an anomaly detection model fitted on $\mathcal{D}_{\text{in}}$.

### 3.1 ANOMALY DETECTION TO SOLVE THE DOMAIN MEMBERSHIP PROBLEM

Detecting whether an instance $x$ is an in-domain instance is equivalent to solving the following problem: *Given task data $\mathcal{T}$ and a sentence $s$, determine if $s$ is likely to come from the distribution generating $\mathcal{T}$ or if $s$ is an anomaly.*

This view helps us make use of a wide variety of anomaly detection techniques developed in literature (Noble & Cook, 2003; Chandola et al., 2009; Chalapathy & Chawla, 2019). To make use of these techniques, we first need a good numeric representation (embedding) with domain discrimination property. We use pre-trained BERT to embed each sentence into a 768 dimensional vector. Once the data is embedded, we need to decide which among the many anomaly detection algorithms proposed in literature should be applied on the embeddings. To decide the anomaly detection method, we propose an evaluation method ranking the techniques based on their discriminative properties.

**Ranking anomaly detection algorithms:** The idea is to treat the anomaly score as the prediction of a classifier distinguishing between in-domain and out-of-domain data. By doing so, we can consider classification metrics such as the f1_score as the score used to rank the anomaly detection algorithm. To do this, we split the in-domain data (the task data) into $\mathcal{D}_{\text{in}}^{\text{train}}, \mathcal{D}_{\text{in}}^{\text{test}}$ using a 90/10 split. We also create out-of-domain data $\mathcal{D}_{\text{out}}$ as a random subset of $\mathcal{D}$ of the same size as $\mathcal{D}_{\text{in}}^{\text{test}}$. We train an anomaly detection algorithm $A$ with $\mathcal{D}_{\text{in}}^{\text{train}}$, and evaluate it's f1_score on the labeled test set composed of the union $\mathcal{D}_{\text{in}}^{\text{test}} \cup \mathcal{D}_{\text{out}}$, where the labels indicate which set the instance originated from. Note that anomaly detection algorithms considered do not require labeled samples for training. Thus, mixing data from $\mathcal{D}_{out}$ does not add much value.

Table 1 provides the results of this evaluation on six anomaly detection algorithms. Details of the tasks can be found in Section 4. We can see that Isolation Forest consistently performs well for most of the tasks. Local Outlier Factor performs almost equally well but is slower in prediction. Although it is possible to adaptively choose for every task the anomaly detection algorithm maximizing the f1_score, we chose to use a single algorithm, Isolation Forests, for the sake of having a simpler technique and generalizable results.

| Task | RC | kNN | PCA | OCS | LOF | IF | Task | RC | kNN | PCA | OCS | LOF | IF |
|---|---|---|---|---|---|---|---|---|---|---|---|---|---|
| CHEMPROT | 0.89 | 0.85 | 0.92 | 0.87 | 0.92 | **0.96** | IMDB | 0.88 | 0.96 | 0.87 | 0.81 | **0.96** | 0.94 |
| ACL-ARC | 0.77 | 0.88 | 0.90 | 0.89 | **0.91** | 0.88 | SCIERC | 0.78 | 0.84 | 0.86 | 0.76 | 0.88 | **0.92** |
| HYPERPARTISAN | 0.86 | 0.86 | 0.95 | 0.98 | 0.91 | **0.98** | HELPFULNESS | 0.82 | 0.89 | 0.83 | 0.76 | 0.83 | **0.92** |
| RCT20K | 0.85 | 0.88 | 0.82 | 0.76 | 0.87 | **0.93** | IMDB | 0.84 | 0.89 | 0.80 | 0.73 | **0.92** | 0.87 |

Table 1: Scores of different anomaly detection algorithms for different tasks. RC: Robust Covariance (Nguyen & Welsch, 2010), kNN: Nearest neighbor (Gu et al., 2019), PCA: Principal Component Analysis (Harrou et al., 2015), OCS: One Class SVM (Schölkopf et al., 2000), LOF: Local Outlier Factor (Breunig et al., 2000), IF: Isolation Forest (Liu et al., 2008)

**Isolation Forest (Liu et al., 2008):** For completeness, we provide a brief description of the Isolation Forest algorithm. Isolation Forest is an unsupervised decision tree ensemble method that identifies anomalies by isolating outliers of the data. It isolates anomalies in data points instead of profiling the normal points. Algorithm works by recursively partitioning the data using a random split between the minimum and maximum value of a random feature. It works due to the observation that outliers are less frequent than the normal points and lie further away from normal points in the feature space. Thus, in a random partitioning, anomalous points would require fewer splits on features resulting in shorter paths and distinguishing from the rest of the points. Anomaly score of a point $x$ is defined as $s(x, n) = 2^{-\frac{E(h(x))}{c(n)}}$, where $E[h(x)]$ is the expected path length of $x$ in various decision trees, $c(n) = 2H(n-1) - 2(n-1)/n$ is the average path length of unsuccessful search in a Binary Tree and $H(n-1)$ is the $n-1$-th harmonic number and $n$ is the number of external nodes.

Now that we chose the anomaly detection technique, we move to discuss the effectiveness of the algorithm in (i) identifying the domain from the task data (ii) identifying the domain related data from the general domain corpus.

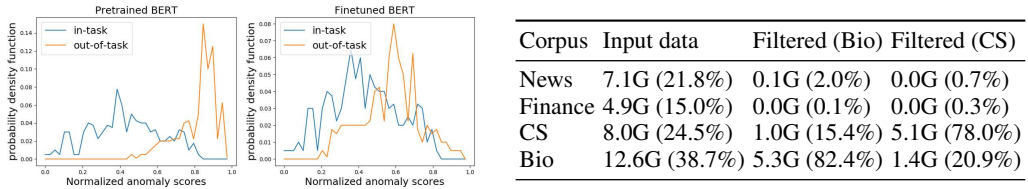

| Corpus | Input data | Filtered (Bio) | Filtered (CS) |
|---|---|---|---|
| News | 7.1G (21.8%) | 0.1G (2.0%) | 0.0G (0.7%) |
| Finance | 4.9G (15.0%) | 0.0G (0.1%) | 0.0G (0.3%) |
| CS | 8.0G (24.5%) | 1.0G (15.4%) | 5.1G (78.0%) |
| Bio | 12.6G (38.7%) | 5.3G (82.4%) | 1.4G (20.9%) |

Figure 3: Sentence anomaly scores for SST with anomaly detection algorithm trained on embeddings from Left: pre-trained BERT, Right: finetuned BERT. In-task: sentences from the task data, out-of-task: sentences from general domain corpus.

Figure 4: Filtering algorithm trained with Bio Abstracts and CS task data. We mix four corpora, filter out 80% of the data and retain the remaining 20% in both cases.

Figure 3 (left) shows that the anomaly detection algorithm is able to distinguish between the in-task-domain data and the out-of-task domain data. These experiments are done for the Sentiment Analysis task (SST) discussed in Section 4. Interestingly, we noticed in our experiments that a language model pre-trained on a diverse corpus is a better choice when compared to a model finetuned on the target domain. We conjecture that the reason is that a finetuned BERT is overly focused on the variations in the task data which are useful for task prediction and forgets information pertaining to different domains which is useful for domain discrimination. We exhibit this phenomenon more clearly in Figure 3 (right) where it is evident that the discriminating ability of the finetuned model is worse.

In order to assess the ability of our model to identify related text we perform the following experiment. First, we create a diverse corpus by taking the union of 4 datasets: News, Finance, CS abstracts and Biology abstracts. Figure 4, column 'Input data' contains their respective sizes. We then train two anomaly score based discriminators, one on CS task data and the other on Bio abstracts. For each

model we choose a threshold that would filter out 80% of the data, and observe the data eventually retained by it. The fraction of data retained from each corpus for each model is given in Figure 4, columns 'Filtered (Bio)' and 'Filtered (CS)'. We see that data from the News and Finance corpus is almost completely filtered as it is quite different than the text in abstracts of academic papers. We also see that a non-negligible percent of the filtered data for the Bio model comes from CS and vice versa. Since both corpora are abstracts of academic papers it makes sense that each corpus contains relevant data for the other. The details related to these corpora are given in Appendix B.

### 3.2 FROM ANOMALY DETECTION SCORES TO DOMAIN ADAPTED PRE-TRAINING

Once the anomaly detection object is trained, we use it to compute the relevance weights i.e. compute $\lambda$ values defined in equation 1. Let the sentences in the pre-training corpus be $s_1, \ldots, s_N$ with anomaly scores $\{A(s_1), \ldots, A(s_N)\}$. We explore two different strategies of $\lambda$ value computation. First is when we normalize and transform the scores to compute continuous values and second when we use threshold and compute $0/1$ values.

**Continuous $\lambda$ values:** We start by normalizing the anomaly scores to be mean zero and variance 1. Let $\mu = (\sum_{i=1}^{N} A(s_i))/N, \sigma = \sqrt{(\sum_{i=1}^{N}(A(s_i) - \mu)^2)/N}$. Then, for every $i \in \{1, \ldots, N\}$, normalized score is $\bar{A}(s_i) = (A(s_i) - \mu)/\sigma$. Using these normalized sentence anomaly scores, we compute the relevance weights as follows: $\lambda(s_i) = \frac{1}{1+e^{-C(\alpha - \bar{A}(s_i))}}$ where $C$ and $\alpha$ are hyper-parameters. $C$ controls the sensitivity of the weight in terms of anomaly score and $\alpha$ controls the fraction of target domain data present in the general domain corpus. $C \to \infty$ results in $0/1$ weights corresponding to discrete $\lambda$ setting whereas $C = 0$ results in no task adaptation setting.

**Discrete $\lambda$ values:** We sort the sentences as per anomaly scores, $A(s_{\sigma(1)}) \leq A(s_{\sigma(2)}) \leq \cdots \leq A(s_{\sigma(N)})$ and pick $\beta$ fraction of the sentences with lowest anomaly scores, $\lambda(s_{\sigma(i)}) = 1$ for $i \in \{1, \ldots, \beta N\}$ and 0 otherwise. Even though this approach is less general than the continuous $\lambda$ values case, it has an advantage of being model independent. We can filter out text, save it and use it to train any language model in a black box fashion. It does not require any change in pre-training or finetuning procedure. However, to utilize this option we need to make a change. Instead of filtering out sentences we need to filter out segments containing several consecutive sentences.

To understand why, suppose we filter out sentence 1 and sentence 10 and none of the sentences in between. When we save the text and construct input instances from it for a language model, then an input instance may contain the end of sentence 1 and the start of sentence 10. This is problematic as sentence 1 and sentence 10 were not adjacent to each other in the original corpus and hence, language model does not apply to them. It distorts the training procedure resulting in worse language models. To resolve this issue, we group sentences into segments and classify the relevance of each segment. Formally, let $\gamma$ be a hyper-parameter and for all $j \in 1, \ldots, \lfloor N/\gamma \rfloor$ let the segment score be $y_j = \sum_{i=(j-1)*\gamma+1}^{j*\gamma} \frac{A(s_i)}{\gamma}$. We sort the segments according to their anomaly scores, $y_{\sigma'(1)} \leq \cdots \leq y_{\sigma'(N/\gamma)}$ and select the $\beta$ fraction with lowest anomaly scores; save the sentences corresponding to these segments. To completely avoid the issue, we may set segment length very large. However, this is not feasible as the diverse nature of pre-training corpus makes sure that large enough segments rarely belong to a specific domain, meaning that the extracted data will no longer represent our target domain. We experimented with a handful of options for the segment length, and found the results to be stable when choosing segments of 15 sentences.

*Continued pre-training instead of pre-training from scratch:* Once we have computed the relevance weights $\lambda(s_i)$, we do not start pre-training the language model from scratch as this is not feasible for each new task/domain. Instead, we start with a language model pre-trained on the general domain corpus and perform additional pre-training for relatively fewer steps with the weighted loss function. In our case, we start with a BERT language model pre-trained for one million steps and continued pre-training with updated loss function for either $50,000$ or $100,000$ steps.

## 4 EXPERIMENTS

We use datasets listed in Table 2 along with a general domain corpus consisting of 8GB of text from Wikipedia articles. We use BERT$_{\text{BASE}}$ model provided in the GluonNLP library for all our

| Task | Train | Dev | Test | C | Task | Train | Dev | Test | C |
|---|---|---|---|---|---|---|---|---|---|
| HYPERPARTISAN | 516 | 64 | 65 | 2 | IMDB | 20000 | 5000 | 25000 | 2 |
| ACL-ARC | 1688 | 114 | 139 | 6 | SST | 67349 | 872 | 1821 | 2 |
| SCIERC | 3219 | 455 | 974 | 7 | AGNEWS | 115000 | 5000 | 7600 | 4 |
| CHEMPROT | 4169 | 2427 | 3469 | 13 | HELPFULNESS | 115251 | 5000 | 25000 | 2 |
| | | | | | RCT20K | 180040 | 30212 | 30135 | 5 |

Table 2: Specification of task datasets. $C$ refers to the number of classes. CHEMPROT (Kringelum et al., 2016) and RCT20K (Dernoncourt & Lee, 2017) are from biomedical domain. HYPERPARTISAN (Kiesel et al., 2019) and AGNEWS (Zhang et al., 2015) are from news domain. HELPFULNESS (McAuley et al., 2015) and IMDB (Maas et al., 2011) are from reviews domain. ACL-ARC (Jurgens et al., 2018) and SCIERC (Luan et al., 2018) are from CS domain. SST (Socher et al., 2013) is a general domain sentiment analysis task.

experiments. It has 12 layers, 768 hidden dimensions per token, 12 attention heads and a total of 110 million parameters. It is pre-trained with a sum of two objectives. First is the masked language model objective where model learns to predict masked tokens. Second is the next sentence prediction objective where sentence learns to predict if sentence $B$ follows sentence $A$ or not. We use learning rate of $0.0001$, batch size 256 and warm-up ratio $0.01$. For finetuning, we pass the final layer [CLS] token embedding through a task specific feed-forward layer for prediction. We use learning rate 3e-5, batch size 8, warm-up ratio $0.1$ and finetune the network for five epochs. In all the experiments, we start with a BERT pre-trained for one million steps and continue pre-training for additional $50,000$ steps in case of discrete $\lambda$, and $100,000$ steps in case of continuous $\lambda$. Also, as mentioned in Section 3.2, we filter out segments instead of sentences and save them. We set the segment length to be 15 sentences and filter out $20\%$ of the data. Pseudo-code of the end-to-end algorithm can be found in Appendix B.

### 4.1 BASELINE METHODS

For each baseline data selection method, we start with a BERT pre-trained on general domain corpus for one million steps as in case of TADPOLE. Then, we continue pre-training the baseline method for the same number of steps as in case of our method. In case of baseline methods which filter general domain corpus, we filter the same fraction of text as in case of our method.

**General:** *Continued pre-training on general domain corpus.* We know that in general longer pre-training leads to a better model. To estimate the impact of extra pre-training, we consider a baseline where we continue pre-training on the general domain corpus.

**Random:** *Continued pre-training on random subset of general domain corpus.*

**Task (Gururangan et al., 2020):** *Continued pre-training on task data.* We continue pre-training on the task data. Since task data is small, we can not pre-train on the task data for as many steps as in other cases. Instead we do 100 epochs, save the model after every 10 epoch and pick the best one.

**LM (Moore & Lewis, 2010):** *Continued pre-training on text filtered via language models trained on task data.* We train two language models, one on the task data and another on a subset of general domain corpus (same size as the task data). We select sentences with lowest scores given by the function $f(s) = H_I(s) - H_O(s)$, where $H_I(s)$ and $H_O(s)$ are the cross-entropy between the $n$-gram distribution and the language model distribution. More formally, cross entropy of a string $s$ with empirical $n$-gram distribution $p$ given a language model $q_I$ is $H_I(s) = -\sum_x p(x) \log q_I(x)$.

**Distance (Wang et al., 2017a):** *Continued pre-training on data filtered via Euclidean distance scoring function.* For each sentence $f$, we consider BERT embedding $v_f$ and compute vector centers $C_{F_{\text{in}}}$ and $C_{F_{\text{out}}}$ of the task data $F_{\text{in}}$ and a random subset of general domain corpus $F_{\text{out}}$; $C_{F_{\text{in}}} = \frac{\sum_{f \in F_{\text{in}}} v_f}{|F_{\text{in}}|}, C_{F_{\text{out}}} = \frac{\sum_{f \in F_{\text{out}}} v_f}{|F_{\text{out}}|}$. We score a sentence $f$ as per the scoring function: $\delta_f = d(v_f, C_{F_{\text{in}}}) - d(v_f, C_{F_{\text{out}}})$. We pick the text with lowest scores.

### 4.2 RESULTS

Table 3 shows the effectiveness of TADPOLE, automatically adapting pre-training to the task domain. Continuing pre-training on the unfiltered corpus (General or Random subset) yields an average gain

| Task | Base | General | Random | Task | LM | Distance | TADPOLE |
|---|---|---|---|---|---|---|---|
| | | Small datasets: # training examples < 5K | | | | | |
| HPRPARTISAN | $70.57_{3.04}$ | $70.97_{2.03}$ | $71.04_{2.32}$ | $70.88_{2.63}$ | $71.47_{2.56}$ | $72.16_{2.14}$ | $\mathbf{73.58}_{2.39}$ |
| ACL-ARC | $72.31_{4.7}$ | $72.38_{3.93}$ | $72.42_{3.71}$ | $72.46_{3.48}$ | $72.40_{1.85}$ | $72.47_{2.64}$ | $\mathbf{72.81}_{3.83}$ |
| SCIERC | $82.84_{1.39}$ | $82.85_{1.38}$ | $82.81_{1.13}$ | $83.18_{1.09}$ | $82.99_{2.75}$ | $83.40_{2.17}$ | $\mathbf{85.85}_{0.95}$ |
| CHEMPROT | $81.62_{0.74}$ | $81.59_{0.67}$ | $81.62_{0.71}$ | $81.63_{0.82}$ | $81.83_{0.74}$ | $81.64_{0.76}$ | $\mathbf{82.41}_{0.62}$ |
| Average Gain | - | $0.11_{0.05}$ | $0.15_{0.05}$ | $0.20_{0.04}$ | $0.34_{0.08}$ | $0.34_{0.06}$ | $\mathbf{1.82}_{0.3}$ |
| | | Large datasets: # training examples $\geq$ 5K | | | | | |
| IMDB | $88.65_{0.24}$ | $88.53_{0.27}$ | $88.63_{0.26}$ | $88.77_{0.39}$ | $88.67_{0.44}$ | $88.69_{0.47}$ | $\mathbf{89.29}_{0.22}$ |
| SST | $92.02_{0.29}$ | $92.21_{0.31}$ | $92.14_{0.24}$ | $92.21_{0.24}$ | $92.25_{0.4}$ | $92.15_{0.35}$ | $\mathbf{92.42}_{0.32}$ |
| AGNEWS | $93.99_{0.13}$ | $94.06_{0.19}$ | $\mathbf{94.09}_{0.11}$ | $94.04_{0.08}$ | $94.03_{0.13}$ | $94.04_{0.11}$ | $94.03_{0.16}$ |
| HELPFULNESS | $69.30_{0.60}$ | $69.39_{0.78}$ | $69.34_{0.58}$ | $69.41_{0.50}$ | $69.58_{0.59}$ | $69.42_{0.69}$ | $\mathbf{69.70}_{0.92}$ |
| RCT20K | $87.52_{0.16}$ | $87.57_{0.16}$ | $87.54_{0.17}$ | $87.60_{0.18}$ | $\mathbf{87.85}_{0.23}$ | $87.62_{0.24}$ | $87.82_{0.13}$ |
| Average Gain | - | $0.02_{0.02}$ | $0.04_{0.01}$ | $0.11_{0.01}$ | $0.18_{0.03}$ | $0.09_{0.01}$ | $\mathbf{0.36}_{0.04}$ |
| | | All datasets | | | | | |
| Average Gain | - | $0.08_{0.05}$ | $0.09_{0.05}$ | $0.15_{0.04}$ | $0.25_{0.08}$ | $0.32_{0.17}$ | $\mathbf{1.01}_{0.36}$ |

Table 3: Performance of TADPOLE and five Baseline methods. Base corresponds to the pre-trained model on general domain corpus with no further pre-training. Baseline methods are mentioned in previous subsection. TADPOLE corresponds to our method with discrete relevance weights. Keeping in line with the previous works, we use the following metrics: accuracy for SST, micro f1 score for CHEMPROT and RCT20K, macro f1 score for ACL-ARC, SCIERC, HELPFULNESS, HPRPARTISAN, IMDB, and AGNEWS. Each model is finetuned eight times with different seeds and the mean value is reported. Subscript correspond to the standard deviation in the finetuned model performance. Average gain corresponds to the average improvement over Base for each of the baseline methods and TADPOLE. Subscript in Average Gain corresponds to the standard deviation in the estimate of the average gain.

less than 0.1%. Adapting pre-training by training on the task data only yields an average gain of 0.15%. Applying popular data selection methods known for domain adaptation including Language Model based relevance score or Distance based relevance score yields a maximum gain of 0.32%. TADPOLE beats all these methods and achieve an average gain of 1.01%. For four tasks with small number of labeled examples (less than 5k), we get a much higher gain of 1.82%. This shows the efficacy of domain adapted finetuning if the task dataset is small. For eight tasks from four domains (all except SST), TADPOLE achieves an average gain of 1.07% whereas the best baseline method achieves an average gain of 0.34%. Models pre-trained on each of the four domain specific corpus can achieve a higher gain (3.37%) over the base model. However, unlike these models, our method has the advantage that it does not require access to any large domain specific corpus. Instead we only need a small task dataset available for finetuning. So, it is applicable to any new task from any new domain. We can also observe in Table 3, that TADPOLE beats all the baseline methods in seven of the nine cases. For RCT20K, it achieves a performance gain (0.3%) on par with the best baseline method (0.33%). For AGNews, gain is insignificant for all the methods as well as models trained on domain specific corpus. We observe this correlation for other tasks as well. Performance boost is higher if the corresponding boost via additional pre-training on large domain specific corpus is higher. Results for this comparison can be found in Appendix E. In Table 3, results are presented for the discrete relevant weight case as they are better when the number of steps available to continue pre-training are small. Results for continuous weights case can be found in Appendix D.

## 5 CONCLUSION

Domain shift in finetuning from Pre-training can significantly impact the performance of deep learning models. We address this issue in the most reasonable setting when we only have access to the labeled task data for finetuning. We adapt data selection methods from Domain Adaptation to adapt pre-training for the downstream task. The existing methods either require sufficiently large task data, or are based on adhoc techniques that do not generalize well across tasks. Our major contribution is providing a new data selection technique that performs well even with very little task data, and generalizes well across tasks.

APPENDIX

## A  RELATED WORK

**Domain Adaptation:** A typical set up for Domain Adaptation involves access to labeled data in source domain, very limited or no labeled data in the target domain and unlabeled data in both source and target domains. This is somewhat different than the setup for our paper where we have access to labeled data with no additional unlabeled data in the task domain and our objective is optimize performance for the same domain. Nevertheless, several techniques of Domain Adaptation have similarities or core components useful for our setup. There are two sets of approaches addressing Domain Adaptation problem: model-centric and data-centric. Model-centric approaches redesign parts of the model: the feature space, the loss function or regularization and the structure of the model (Blitzer et al., 2006; Pan et al., 2010; Ganin et al., 2016). A recent such approach, appropriate for our setting is called Pivot-based Domain Adaptation; it has recently been applied to Task Adaptive Pre-training when there is additional unlabeled task data available Ben-David et al. (2020). In a nutshell, the idea is to distinguish between pivot and non-pivot features, where pivot features behave similarly in both domains. Then, by converting the non-pivot to pivot features, one can make use of a model trained on the source data. This approach does not work well when the target data is small since the mapping of non-pivot to pivot features cannot be trained with a limited size dataset. Since our technique is data-centric and applies to the regime of a small target corpus, we do not further analyze this or any other model-centric approach.

Data-centric approaches for domain adaptation include pseudo-labeling, using auxiliary tasks and data selection. Pseudo-labeling apply a trained classifier to predict labels on unlabeled instances which are then treated as 'pseudo' gold labels for further training (Abney, 2007; Cui & Bollegala, 2019). Auxiliary-task domain adaptation use labeled data from auxiliary tasks via multi-task learning (Peng & Dredze, 2016) or intermediate-task transfer (Phang et al., 2018; 2020). The methods most relevant to us are those of data selection and are discussed above in detail.

## B  DATASETS IN ACCURACY ESTIMATION OF ANOMALY SCORE BASED DATA FILTRATION

**CS task data:** To train anomaly score discriminator for CS data, we use the tasks data from ACL-ARC and SCIERC. Details of these datasets are mentioned in Section 4.

**CS and Bio Abstracts:** Semantic Scholar corpus (Ammar et al., 2018) contains datasets from a variety of domain. We filter out text based on the domain field and only keep the abstracts from CS and bio domain.

**News:** We use REALNEWS (Zellers et al., 2019) corpus containing news articles from 500 news domains indexed by Google News. It is obtained by scraping dumps from Common Crawl.

**Finance:** We use the TRC2-financial dataset. This a subset of Reuters TRC24 corpus containing news articles published between 2008 and 2010. It can be obtained by applying here: https://trec.nist.gov/data/reuters/reuters.html

## C  PSEUDO CODE

Algorithm 1 shows the pseudo code for the case of continuous relevance weights. Discrete relevance weight setting is same as $C \to \infty$. As discussed in 3.2, in case of discrete relevance weights, we filter out segments containing several consecutive sentences. We experimented with several options for the segment length and found the stable segment length to be $15$ sentences. Here, a sentence is a consecutive piece of text such that when applied through the BERT tokenizer, it results in 256 sentences.

**Algorithm 1** Task Adaptive Pre-training

Input: Pre-trained model $B$, Pre-training instances $x_1,\ldots,x_N$, task data $\mathcal{T}$, $(C,\alpha)$, #steps
**Stage 1: Instance weight computation**
Let the sentences of the task be $s_1,\ldots,s_t$ with sentence embeddings $P = \{\text{Embed}(s_1),\ldots,\text{Embed}(s_t)\}$.
Let a random subset of pre-training instances (sentences of these instances) be $s'_1,\ldots,s'_{t/10}$ with BERT based sentence embeddings $N = \{\text{Embed}(s'_1),\ldots,\text{Embed}(s'_{t/10})\}$
Train an anomaly detection object, $\text{IF} = \text{IsolationForest}(P \cup N)$
For $i \in [N]$, let $S(x_i) = \text{IF.score}(\text{Embed}(x_i))$
Let $\mu = \frac{1}{N}\sum_{i=1}^{N} S(x_i)$ and $\sigma = \sqrt{\frac{1}{N}\sum_{i=1}^{N}(S(x_i)-\mu)^2}$.
For every $i \in [N]$, $\bar{S}(x_i) = \frac{S(x_i)-\mu}{\sigma}$.
For every $i \in [N]$, $\lambda(x_i) = \frac{1}{1+e^{-C(\alpha-\bar{S}(x_i))}}$
**Stage 2: Adaptation of pre-training to target domain**
Continue training language model $B$ for #steps on instances $x_1,\ldots,x_N$ with instance weights $\lambda(x_1),\ldots,\lambda(x_N)$.
Finetune resulting model on the labeled task data $\mathcal{T}$

| Task | Base | Discrete | Continuous-1 | Continuous-3 |
|---|---|---|---|---|
| CHEMPROT | $81.62_{0.74}$ | $\mathbf{82.41}_{0.62}$ | $81.74_{0.81}$ | $81.64_{0.83}$ |
| RCT20K | $87.52_{0.16}$ | $\mathbf{87.82}_{0.13}$ | $87.49_{0.28}$ | $87.56_{0.22}$ |
| HPRPARTISAN | $70.57_{3.04}$ | $\mathbf{73.58}_{2.39}$ | $70.94_{1.98}$ | $71.29_{2.95}$ |
| AGNEWS | $93.99_{0.13}$ | $\mathbf{94.03}_{0.16}$ | $94.01_{0.14}$ | $94.01_{0.15}$ |
| HELPFULNESS | $69.30_{0.60}$ | $\mathbf{69.70}_{0.92}$ | $69.35_{0.5}$ | $69.37_{0.44}$ |
| IMDB | $88.65_{0.24}$ | $\mathbf{89.29}_{0.22}$ | $88.63_{0.51}$ | $88.71_{0.46}$ |
| ACL-ARC | $72.31_{4.7}$ | $\mathbf{72.81}_{3.83}$ | $72.26_{2.33}$ | $72.36_{2.12}$ |
| SCIERC | $82.84_{1.39}$ | $\mathbf{85.85}_{0.95}$ | $83.14_{1.96}$ | $83.13_{2.65}$ |
| SST | $92.02_{0.29}$ | $\mathbf{92.42}_{0.32}$ | $92.11_{0.32}$ | $92.13_{0.37}$ |

Table 4: Comparison of discrete vs continuous relevance weight setting. Base corresponds to the pre-trained model on general domain corpus with no further pre-training. Discrete refers to TADPOLE with discrete relevance weights/filtered out text and pre-trained additionally for 50000 steps. Continuous-x refers to TADPOLE with continuous relevance weights and pre-trained additionally for $x*100,000$ more steps. Metrics used for different tasks: accuracy for SST, micro f1 score for CHEMPROT and RCT20K, macro f1 score for ACL-ARC, SCIERC, HELPFULNESS, HPRPARTISAN, IMDB, and AGNEWS. Each model is finetuned eight times with different seeds and the mean value is reported. Subscript correspond to the standard deviation in the finetuned model performance.

## D CONTINUOUS RELEVANCE WEIGHTS

We see in Table 4 that a model additionally pre-trained for 50,000 with discrete $\lambda$ values consistently over performs the continuous case even when we train with continuous relevance weights for far higher number of steps. This is because of the fact that many of those steps yield virtually no training at all. For instance, suppose the relevance weights are uniformly distributed between 0 and 1; $[0.1, 0.2, 0.3, 0.4, 0.5, 0.6, 0.7, 0.8, 0.9, 1]$. Then, in discrete case we pick the top two sentences and thus two steps are sufficient to train on these most relevant sentences (assume batch size is 1). However, in continuous case, we need to train the model for ten steps to train on these top two relevant sentences. Thus, we need many more steps to achieve and beat the performance achieved in the Discrete case. An open question is to combine the two settings so as to benefit from the generality of Continuous case and efficiency of the discrete case.

## E PERFORMANCE BOOST WITH DOMAIN-SPECIFIC CORPUS VS TADPOLE

We compare the performance boost we achieved due to TADPOLE with the performance boost we achieve if we have access to large pre-training corpus. In Table 5, we list the gain in performance in both cases over eight tasks from four domains. We see that the performance boost is higher with

| Task | TADPOLE | Domain Corpus |
|---|---|---|
| CHEMPROT | 0.79 | 2.3 |
| RCT20K | 0.3 | 0.4 |
| HYPERPARTISAN | 3.01 | 1.6 |
| AGNEWS | 0.04 | 0.0 |
| HELPFULNESS | 0.4 | 1.4 |
| IMDB | 0.64 | 5.4 |
| ACL-ARC | 0.5 | 3.5 |
| SCIERC | 3.01 | 12.4 |

Table 5: Performance boost via TADPOLE vs pre-training on domain specific corpus. TADPOLE corresponds to our method with discrete relevance weights/filtered out text and pre-trained additionally for 50000 steps. Domain Corpus refers to the model trained in Gururangan et al. (2020) over the domain same as the downstream task. Metrics used for different tasks: accuracy for SST, micro f1 score for CHEMPROT and RCT20K, macro f1 score for ACL-ARC, SCIERC, HELPFULNESS, HPRPARTISAN, IMDB, and AGNEWS. Each model is finetuned eight times with different seeds. We report the difference in the mean value of performance between the model with additional pre-training and base model with no additional pre-training.

TADPOLE if the corresponding boost is higher with domain specific corpus. Thus if there is a large domain shift between the general domain corpus and the task data, as can be measured by the performance boost via large pre-training corpus, then TADPOLE is able to achieve large performance boost via Task Adaptation. Scale of numbers in the two columns are not directly comparable due to the following two reasons. First is that additional pre-training done is Gururangan et al. (2020) is for almost as many steps as the number of steps required to pre-train a network from scratch. However, in our case additional pre-training is done for only $5\%$ of the number of steps required to pre-train a network from scratch. Second reason is that the model used in (Gururangan et al., 2020) is different, ROBERTA. Also, the general domain corpus is different and thus the domain shift is not exactly the same as in our case. The point however remains the same, which is that as the target domain is further away from the pre-training corpus, the benefits of TADPOLE increase.

# HANDLING LONG-TAIL QUERIES WITH SLICE-AWARE CONVERSATIONAL SYSTEMS

**Cheng Wang, Sun Kim, Taiwoo Park, Sajal Choudhary**
Amazon Alexa AI
{`cwngam, kimzs, parktaiw, sajalc`}@amazon.com

**Sunghyun Park, Young-Bum Kim, Ruhi Sarikaya, Sungjin Lee**
Amazon Alexa AI
{`sunghyu, youngbum, rsarikay, sungjinl`}@amazon.com

## ABSTRACT

We have been witnessing the usefulness of conversational AI systems such as Siri and Alexa, directly impacting our daily lives. These systems normally rely on machine learning models evolving over time to provide quality user experience. However, the development and improvement of the models are challenging because they need to support both high (head) and low (tail) usage scenarios, requiring fine-grained modeling strategies for specific data subsets or slices. In this paper, we explore the recent concept of slice-based learning (SBL) (Chen et al., 2019) to improve our baseline conversational skill routing system on the tail yet critical query traffic. We first define a set of labeling functions to generate weak supervision data for the tail intents. We then extend the baseline model towards a slice-aware architecture, which monitors and improves the model performance on the selected tail intents. Applied to de-identified live traffic from a commercial conversational AI system, our experiments show that the slice-aware model is beneficial in improving model performance for the tail intents while maintaining the overall performance.

## 1 INTRODUCTION

Conversational AI systems such as Google Assistant, Amazon Alexa, Apple Siri and Microsoft Cortana have become more prevalent in recent years (Sarikaya, 2017). One of the key techniques in those systems is to employ machine learning (ML) models to route a user's spoken utterance to the most appropriate skill that can fulfill the request. This requires the models to first capture the semantic meaning of the request, which typically involves assigning the utterance query to the candidate domain, intent, and slots (El-Kahky et al., 2014). For example, "Play Frozen" can be interpreted with *Music* as the domain, *Play Music* as the intent, and *Album Name:Frozen* as the slot key and value. Then, the models can route the request to a specific skill, which is an application that actually executes to deliver an experience (Li et al., 2021). For commercial conversational AI systems, there usually exists a large-scale dataset of user requests with ground-truth semantic interpretations and skills (e.g., through manual annotations and hand-crafted rules or heuristics). Along with various contextual signals, it is possible to train ML models (e.g., deep neural networks) with high predictive accuracy in routing a user request to the most appropriate skill, which then can continue to optimize towards better user experience through implicit or explicit user feedback (Park et al., 2020).

Nevertheless, developing such ML models or improving existing ones towards better user experience is still challenging. One hurdle is the imbalance in the distribution of the user queries with a long tail in terms of traffic volume. This often makes it difficult for the ML models to learn the patterns from the long-tail queries, some of which could be for critical features. Several approaches have been proposed to address such imbalance issue (Smith et al., 2014; He et al., 2008; Chawla et al., 2002). However, they are mainly based on applying reverse-discriminative sampling strategies,

for example, over-sampling minority and/or under-sampling majority. The sampling methods are usually insufficient in inspecting and improving model performance on pre-defined data subgroups.

In this work, we focus on the problem of imbalanced queries, specifically on tail but critical intents, in the context of the recently proposed slice-based learning (SBL) (Chen et al., 2019). SBL is a novel programming model that sits on top of ML systems. The approach first inspects particular data subsets (Ratner et al., 2019), which are called slices, and it improves the ML model performance on those slices. While the capability of monitoring specific slices is added to a pre-trained ML model (which is termed the *backbone* model), the approach has shown that overall performance across the whole traffic is comparable to those without SBL. Motivated by this idea, we propose to adopt the SBL concept to our baseline skill routing approach (we term the baseline model as $\mathbf{P}$; please refer to Sec. 3.1 for details) to improve its performance on tail yet critical intent queries while keeping the overall performance intact. First, we define slice functions (i.e., labeling functions) to specify the intents that we want to monitor. A pre-trained $\mathbf{P}$ is used as a backbone model for extracting the representation for each query. Then, we extend $\mathbf{P}$ to a slice-aware architecture, which learns to attend to the tail intent slices of interest.

We perform two experiments using a large-scale dataset with de-identified customer queries. First, we examine the attention mechanism in the extended model $\mathbf{P}$ with SBL. In particular, we test two attention weight functions with different temperature parameters in computing the probability distribution over tail intent slices. Second, we compare SBL to an upsampling method in $\mathbf{P}$ for handling tail intents. Our experiments demonstrate that SBL is able to effectively improve the ML model performance on tail intent slices as compared to the upsampling approach, while maintaining the overall performance.

We describe the related work in Section 2. In Section 3, we explain the baseline skill routing model, $\mathbf{P}$, and then elaborate how to extend it to a slice-aware architecture. The experiment results are reported in Section 4, and in Section 5, we discuss the advantages and potential limitations of applying SBL in our use case. We conclude this work in Section 6.

## 2 RELATED WORK

### 2.1 SLICE-BASED LEARNING

Slice-based learning (SBL) (Chen et al., 2019) is a novel programming model that is proposed to improve ML models on critical data slices without hurting overall performance. A core idea of SBL is to represent a sample differently depending on the data subset or slice to which it belongs. It defines and leverages slice functions, i.e., pre-defined labeling functions, to generate weak supervision data for learning slice-aware representations. For instance, in computer vision (CV) applications, a developer can define object detection functions to detect whether an image contains a bicycle or not. In natural language understanding (NLP) applications, a developer can define intent-specific labeling functions such as for *Play Music* intent. SBL exhibits better performance than a mixture of experts (Jacobs et al., 1991) and multi-task learning (Caruana, 1997), with reduced run-time cost and parameters (Chen et al., 2019). Recently Gustavo et al. (Penha & Hauff, 2020) have employed the concept of SBL to understand failures of ranking models and identify difficult instances in order to improve ranking performance. Our work applies the idea to improve skill routing performance on low traffic but critical intents in conversational AI systems.

### 2.2 WEAKLY SUPERVISED LEARNING

Weakly supervised learning attempts to learn predictive models with noisy and weak supervision data. Typically, there are three types of weak supervision: incomplete supervision, inexact supervision, and inaccurate supervision (Zhou, 2018). Various weakly supervised ML models are developed in NLP (Medlock & Briscoe, 2007; Huang et al., 2014; Wang & Manning, 2014) and in CV (Prest et al., 2011; Oquab et al., 2015; Peyre et al., 2017). Recently, promising approaches have been proposed to generate weak supervision data by programming training data (Ratner et al., 2016). In a large-scale industry setting, weak supervision data are highly desired given that human annotations are costly and time-consuming. Our work relates to weakly supervised learning in terms of inaccurate supervision. We split queries into different groups (slices) by defining labeling functions (slice functions). Each group is assigned with a group identity label. In practice, the slice functions may

not perfectly assign labels to input data as mentioned in SBL (Chen et al., 2019; Cabannnes et al., 2020).

### 2.3 CONVERSATIONAL SKILL ROUTING MODELS

In conversational AI systems, a skill refers to the application that actually executes on a user query or request to deliver an experience, such as playing a song or answering a question. The skills often comprise both first-party and third-party applications (Li et al., 2021). The skill routing is a mechanism that maps users' queries, given contextual information such as semantic interpretations and device types, to an appropriate application. The routing decision is usually determined by an ML model that is separate from typical natural language understanding (NLU) models for domain, intent, slot parsing. Please refer to section 3.1 for more details.

## 3 SLICE-AWARE CONVERSATIONAL SKILL ROUTING MODELS

This section explains our skill routing model (backbone model) and then explains how we extend the backbone model to a slice-aware architecture by adapting the concept from SBL (Chen et al., 2019).

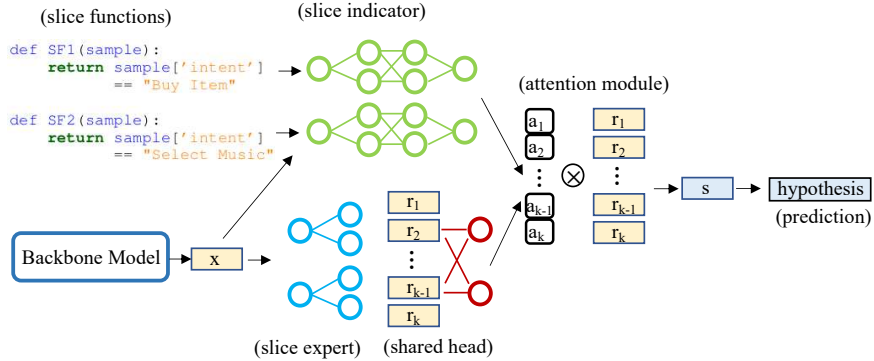

Figure 1: The slice-aware conversational skill routing model architecture for handling low traffic but critical intents. It consists of six components: (1) *slice functions* define tail intent slices that we want to monitor; (2) *backbone model* is our pre-trained skill routing model **P** that is used for feature extraction; (3) *slice indicators* are membership functions to predict if a sample query belongs to a tail slice; (4) *slice experts* aim to learn slice-specific representations; (5) *shared head* is the base task predictive layer across experts; (6) An *attention module* is used to re-weight the slice-specific representations **r** and form a slice-aware representation **s**. Finally, the learned **s** is used to predict a final hypothesis (associated skill). The predicted hypothesis is used to serve a user query.

### 3.1 BACKBONE MODEL

We take our baseline skill routing approach (**P**) as the backbone model and aim to make it a slice-aware architecture. **P** is a skill routing model, which takes in a list of routing candidates to select the most appropriate one. Each routing candidate is represented as a hypothesis with various contextual signals, such as utterance text, device type, semantic interpretation, and associated skill. While some contextual signals are common across all hypotheses, some are unique due to the presence of multiple competing semantic interpretations and skill-specific context. The core component of **P** consists of attention-based bi-directional LSTMs (Hochreiter & Schmidhuber, 1997; Graves & Schmidhuber, 2005) with fully connected layers on top of it. Formally, Let $X$ be the set of query signals (e.g., utterance text, semantic interpretations, device type, etc.), $H = \{h_1, ..., h_n\}$ be the hypothesis list and $h_g \in H, g = [1, n]$ be the ground-truth hypothesis. The learning objective is to minimize the binary cross entropy:

$$\zeta_{base} = \mathcal{L}_{bce}(\pi(\mathcal{M}(X, H)), h_g), \qquad (1)$$

3

```python
def intent_based_slice_function_1(sample):
    return sample['intent'] == "Buy Item"

def intent_based_slice_function_2(sample):
    return sample['intent'] == "Select Music"

def intent_based_slice_function_3(sample):
    return sample['intent'] == "Buy Book"
```

Table 1: The slice functions (SFs) which split user queries into multiple data slices according to the pre-defined tail intents. The non-tail intents are in a base slice. Note SFs are only available at training stage for generating weak supervision labels. At inference stage, SFs will not be applied.

where $\pi$ is a linear predictive layer which outputs a prediction over hypotheses $\hat{H} = \{\hat{h}_1, ..., \hat{h}_n\}$, and $\mathcal{M}$ is a set of multiple neural network layers, which extract the representation $\mathbf{x} \in \mathbb{R}^{n \times d}$ for a given $(X, H)$ pair, i.e., $\mathbf{x} = \mathcal{M}(X, H)$.

To evaluate the effectiveness of trained $\mathbf{P}$, we define offline evaluation metric called replication accuracy (RA):

$$RA(\mathcal{D}_{test}) = \sum_{(X,H,h_g) \in \mathcal{D}_{test}} \frac{\mathbb{I}(\hat{h}_g = h_g)}{|\mathcal{D}_{test}|}. \tag{2}$$

The replication accuracy measures how effectively the trained model $\mathbf{P}$ replicates the current skill routing behavior in production which is a combination of ML model and rules. Though $\mathbf{P}$ achieves high performance, replicating most of heuristic patterns, it suffers from low RA in low-volume traffic, i.e., the tail user queries. We later introduce how we extend $\mathbf{P}$ with a slice-aware component.

### 3.2 SLICE-AWARE ARCHITECTURE

As presented in Figure 1, a slice-aware architecture consists of several components.

**Slice Function**. We first define slice (or labeling) functions to slice user queries according to intent (e.g., "Buy Book"). The selected intents have a small number of query instances, making the model $\mathbf{P}$ difficult to learn data patterns from tail intents. Each sample is assigned a slice label $\gamma \in [0, 1]$ in $\{\gamma_1, \gamma_2, ..., \gamma_k\}$ for supervision. $s_1$ is the base slice, and $s_2$ to $s_k$ are the tail slices.

**Slice Indicator**. For each tail intent slice, a slice indicator (membership function) is learned to indicate whether a sample belongs to this particular slice or not. For a given representation $\mathbf{x} \in \mathbb{R}^{n \times d}$ from the backbone model, we learn $u_i = f_i(\mathbf{x}; \mathbf{w}_i^f)$, $\mathbf{w}_i^f \in \mathcal{R}^{d \times 1}$, $i \in \{1, .., k\}$ that maps $\mathbf{x}$ to $\mathbf{u} = \{u_1, ..., u_k\}$. $f_i$ is trained with $\{\mathbf{x}, \gamma\}$ pairs with the binary cross entropy $\zeta_{ind} = \sum_i^k \mathcal{L}_{bce}(\mathbf{u}_i, \gamma_i)$.

**Slice Expert**. For each tail intent slice, a slice expert $g_i(\mathbf{x}; \mathbf{w}_i^g)$, $\mathbf{w}_i^g \in \mathcal{R}^{d \times d}$ is used to learn a mapping from $\mathbf{x} \in \mathbb{R}^{n \times d}$ to a slice vector $r_i \in \mathcal{R}^d$ with the samples only belonging to the tail slice. Followed by a **shared head**, which is shared across all experts and maps $r_i$ to a prediction $\hat{h} = \varphi(r_i; \mathbf{w}_s)$, $g_i$ and $\varphi$ are learned on the base (original) task with ground-truth label $h_g$ by $\zeta_{exp} = \sum_i^k \gamma_i \mathcal{L}_{bce}(\hat{h}, h_g)$.

**Attention Module**. The attention module decides how to pay special attention to the monitored slices. The distribution over slices (or attention weights) are computed based on stacked $k$ membership likelihood $P \in \mathbb{R}^k$ and stacked $k$ experts' prediction confidence $Q \in \mathbb{R}^{k \times c}$ as described in (Chen et al., 2019):

$$a2 = \text{SOFTMAX}(P + |Q|). \tag{3}$$

Note, the above equation is used when $c = 1$ (i.e., binary classification). As our task is a multi-class classification task where $c \geq 2$, we use an additional linear layer to transform $Q \in \mathbb{R}^{k \times c}$ to $\phi(Q) \in \mathbb{R}^k$. Finally, we experiment with the following different ways to compute attention weights, i.e., slice distribution:

4

$$a1 = \text{SOFTMAX}(P/\tau) \quad (4)$$
$$a2 = \text{SOFTMAX}([P + |\phi(Q)|]/\tau). \quad (5)$$

In Eq. 4, we only use the output of the indicator function (membership likelihood) in computing attention weights. In Eq. 5 we use both the membership likelihood and the transformed experts' prediction scores. The $\tau$ is a temperature parameter. In principle, smaller $\tau$ can lead to a more confident slice distribution (Wang & Niepert, 2019; Wang et al., 2021), hence we aim to examine if a small $\tau$ helps improve the routing performance.

## 4 EXPERIMENTS

We evaluate the skill routing model **P** with slice-based learning (SBL) (Chen et al., 2019) (we term it as **S**) by performing two groups of experiments. First, we test the attention module with different methods of computing the attention weights over slices. Second, we compare the effectiveness of SBL against upsampling – a commonly used method for handling tail data.

### 4.1 EXPERIMENT SETUP AND IMPLEMENTATION DETAILS

We obtained live traffic from a commercial conversational AI system in production and processed the data so that individual users are not identifiable. We randomly sampled to create an adequately large data set for each training and test dataset. We further split the training set into training and validation sets with a ratio of 9:1. We used the replication accuracy (Eq. 2) to measure the model performance.

The existing production model **P** and its extension with SBL were implemented with Pytorch (Paszke et al., 2019). The hidden unit size for slice component was 128. All models were trained on AWS p3.8xlarge instances with Intel Xeon E5-2686 CPUs, 244 GB memory, and 4 NVIDIA Tesla V100 GPUs. We used Adam (Kingma & Ba, 2014) with a learning rate of 0.001 as the optimizer. Each model was trained with 10 epochs with the batch size of 256. We split the user queries into 21 data slices in total, one base slice and the rest for 20 tail intent slices. For each extracted query representation **x** for the tail intents, we add a Gaussian noise $\mathbf{x} = \mathbf{x} + \delta$, $\delta \sim \mathcal{N}(0, 0.005)$ to augment the tail queries.

### 4.2 EXPERIMENTS ON THE ATTENTION MECHANISMS

Table 2 shows the absolute score difference in replication accuracy between the baseline model and its SBL extension, having the baseline model's all-intent accuracy as a reference. As shown in the table, the slice-based approaches maintain the baseline performance overall, but the RA performance is lifted on the monitored tail slices. The best attention mechanism outperforms the baseline by 0.1% in tail intents' replication accuracy[1]. Tuning the temperature parameter between $\tau = 0.1$ or $\tau = 1.0$ does not significantly improve model performance on the tail intents.

| Attention Methods | All Intents (%) | Tail Intents (%) |
|---|---|---|
| **P** (baseline model) | >99 | –1.45 |
| SBL, Eq. (4), $\tau = 1.0$ | +0.01 | –1.35 |
| SBL, Eq. (5), $\tau = 1.0$ | +0.01 | –1.36 |
| SBL, Eq. (4), $\tau = 0.1$ | +0.01 | –1.34 |
| SBL, Eq. (5), $\tau = 0.1$ | +0.01 | –1.38 |

Table 2: The performance comparison of the baseline model **P** and its SBL extension with different attention weights in replication accuracy. All data points denote the absolute difference from the baseline model's all intents accuracy value.

---

[1] Given the large volume of query traffic per day, 0.1% is still a significant improvement in our system.

### 4.3 Comparison between Slice-Based Learning and Upsampling

As upsampling is a widely used method to alleviate the tail data problem, we compare the performance between SBL and upsampling methods. Note SBL offers an additional advantage for inspecting particular tail data groups which are also critical. We denote the models as the following:

- $\mathbf{P}$ is the baseline model that is trained without applying upsampling.
- $\mathbf{S}$ is an extension of $\mathbf{P}$ (as a backbone model) to be a slice-aware model, which is trained with same training set as $\mathbf{P}$.
- $\mathbf{P}_{up}$ is the baseline model that is trained with applying upsampling.
- $\mathbf{S}_{up}$ is an extension of $\mathbf{P}_{up}$ to be a slice-aware model, which is trained with same training set as $\mathbf{P}_{up}$.

All the trained models are evaluated on the same test set. Among the aforementioned attention method choices, Eq. 4 with $\tau = 1.0$ is employed for $\mathbf{S}$ and $\mathbf{S}_{up}$. Our primary goal is to see whether $\mathbf{S}$ can improve $\mathbf{P}_{up}$.

Table 3 shows the performance comparison. When comparing $\mathbf{P}_{up}$ and $\mathbf{S}$, we can see $\mathbf{S}$ achieves slightly better performance for all intents. For the monitored tail intents, $\mathbf{S}$ achieves a slightly higher score as compared to $\mathbf{P}_{up}$.

| Models | All Intents (%) | Tail Intents (%) |
|---|---|---|
| $\mathbf{P}$ | >99 | –1.41 |
| $\mathbf{P}_{up}$ | 0.00 | –1.47 |
| $\mathbf{S}$ | +0.01 | –1.30 |
| $\mathbf{S}_{up}$ | +0.01 | –1.37 |

Table 3: Performance comparison between the baseline model and its slice-aware architecture. $\mathbf{P}$ is the baseline model without upsampling, $\mathbf{P}_{up}$ is $\mathbf{P}$ with upsampling. $\mathbf{S}$ is the slice learning model with $\mathbf{P}$ as the backbone model, and $\mathbf{S}_{up}$ is the slice learning model with $\mathbf{P}_{up}$ as the backbone model. All data points are absolute score difference from the baseline model's all intent accuracy value.

Table 4 presents the absolute RA difference between the baseline and slice-aware models for the monitored 20 tail intents. Comparing $\mathbf{S}$ and $\mathbf{P}_{up}$, $\mathbf{S}$ improves the model performance on 14 tail intents. Compared to $\mathbf{P}_{up}$, $\mathbf{S}$ shows the comparable performance lift while effectively suppressing performance drops, for example, intent IDs 2, 3, 6, 15, and 20. As a result, $\mathbf{P}_{up}$ shows lower performance on 12 intents out of 20 (–2.41% on average), while $\mathbf{S}$ did on only 5 intents (–0.21% on average). This suggests the capability of slice-based learning in treating target intents through the slice-aware representation.

## 5 Discussion

In our experiments, we have shown the effectiveness of SBL in terms of improving model performance on tail intent slices. It is beneficial to have ML models which are slice-aware, particularly when we want to inspect some specific and critical but low-traffic instances. Although the overall performance gain of slice-aware approach compared to the upsampling was marginal, it is worthwhile to note that the slice-aware approach was able to lift up the replication accuracy for more number of tail intents while minimizing unexpected performance degradation that was more noticeable in the upsampling approach. This result implies that the slice-aware approach has more potential in stably and evenly supporting tail intents.

On the other hand, we also note a potential limitation of SBL in the case of addressing tail intents in the industry setting. As we increase the number of tail intents, for instance to 200 intents, the model's complexity increases as well, given that an indicator function and an expert head are needed for each slice. However, this does not necessarily diminish the value of the slice-aware architecture, as the upsampling method offers no chance for us to inspect and analyze model failures on particular slices.

| Tail Intent ID | P | $P_{up}$ | S | $S_{up}$ | Sample Size |
|---|---|---|---|---|---|
| 1 | >99 | –0.03 | 0.00 | –0.02 | Over 10K |
| 2 | >96 | –0.4 | +0.04 | –0.21 | Over 10K |
| 3 | >96 | –0.19 | +0.09 | –0.18 | Over 10K |
| 4 | >72 | +0.07 | +1.98 | +0.96 | Over 10K |
| 5 | >99 | +0.01 | –0.01 | 0.00 | Over 10K |
| 6 | >96 | –0.09 | +0.02 | –0.11 | Over 10K |
| 7 | >96 | +0.03 | +0.02 | –0.04 | Over 10K |
| 8 | >99 | +0.15 | +0.01 | +0.19 | Over 10K |
| 9 | >96 | –0.24 | +0.06 | +0.07 | Over 10K |
| 10 | >96 | +0.08 | –0.03 | 0.00 | Between 1K - 10K |
| 11 | >99 | 0.00 | 0.00 | 0.00 | Between 1K - 10K |
| 12 | >96 | +0.55 | –0.13 | +0.46 | Between 1K - 10K |
| 13 | >96 | +0.36 | +0.42 | +0.53 | Between 1K - 10K |
| 14 | >93 | –0.39 | –0.14 | –0.42 | Between 1K - 10K |
| 15 | >93 | –3.29 | –0.73 | –1.46 | Between 1K - 10K |
| 16 | >96 | –1.2 | 0.00 | –0.93 | Below 1K |
| 17 | >96 | –0.71 | 0.00 | –0.71 | Below 1K |
| 18 | >99 | –0.16 | 0.00 | –0.16 | Below 1K |
| 19 | >99 | –0.96 | 0.00 | –1.15 | Below 1K |
| 20 | >96 | –21.21 | 0.00 | –18.18 | Below 1K |

Table 4: Score (in %) differences in RA between the baseline and slice-aware approaches at the intent level. The baseline model's accuracy scores are rounded down to the nearest multiple of 3 percent, while the other models' are absolute score differences from the baseline ones. We denote each intent with their IDs. Sample Size is the number of random instances used for testing.

Further studies are necessary to employ and fine-tune the slice-based approach to serve tail traffic in a cost-effective way.

## 6 CONCLUSION

In this work, we applied and implemented the concept of slice-based learning to our skill routing model for a large-scale commercial conversational AI system. To enable the existing model to pay extra attention to selected tail intents, we tested different ways of computing slice distribution by using membership likelihood and experts' prediction confidence scores. Our experiments show that the slice-based learning can effectively and evenly improve model performance on tail intents while maintaining overall performance. We also compared the slice learning method against upsampling in terms of handling tail intents. The results suggest that slice-based learning outperforms upsampling by a small margin, while more evenly uplifting tail intents' performance. A potential future work would be to explore how to adapt SBL to monitor a large number of slices with minimum model and runtime complexity.

# TABULAR DATA MODELING VIA CONTEXTUAL EMBEDDINGS

**Xin Huang** *
Amazon AI
xinxh@amazon.com

**Ashish Khetan**
Amazon AI
khetan@amazon.com

**Milan Cvitkovic**
PostEra
mwcvitkovic@gmail.com

**Zohar Karnin**
Amazon AI
zkarnin@amazon.com

## ABSTRACT

We introduce TabTransformer, a new tabular data modeling architecture based on deep self-attention Transformers. Our model works by embedding categorical features in a robust and contextual manual, resulting in better prediction performance. We evaluate TabTransformer for supervised setting through extensive experiments on fifteen publicly available datasets, and conclude that it outperforms the state-of-the-art deep learning methods for tabular data by at least 1.0% on mean AUC. Furthermore, for the semi-supervised setting we develop an unsupervised pre-training and fine-tuning paradigm to learn data-driven contextual embeddings, resulting in an average 2.1% AUC lift over the state-of-the-art methods. Lastly, we demonstrate that the contextual embeddings learned from TabTransformer provide better interpretability, and are highly robust against both missing and noisy data features.

## 1 INTRODUCTION

Tabular data regression and classification are crucial to many real-world applications such as recommender systems (Cheng et al., 2016), online advertising (Song et al., 2019), and sales forecasting (Pavlyshenko, 2019). Many machine learning competitions such as Kaggle (Kaggle, 2020) and KDD Cup (SIGKDD, 2020) are primarily designed to solve problems in tabular domain, where various machine learning models are built to take each instance (row of tabular data) as input and map it to a target value.

The state-of-the-art for modeling tabular data is tree-based ensemble methods such as the gradient boosted decision trees (GBDT) (Chen & Guestrin, 2016). This differs from modeling image and text data where all the existing competitive models are based on deep learning (Sandler et al., 2018; Devlin et al., 2019). The tree-based ensemble models are accurate, fast to train, and easy to interpret, making them highly favourable among machine learning practitioners. However, their limitations are significant compared with deep learning models: (a) they do not allow efficient end-to-end learning of image/text encoders in presence of multi-modality along with tabular data; (b) they do not fit into the state-of-the-art semi-supervised learning framework due to unreliable probability estimation produced by basic decision tree (Tanha et al., 2017); and (c) they do not enjoy the SoTA deep learning methods (Devlin et al., 2019) to handle missing and noisy data features.

A classical and popular model that is trained using gradient descent and hence allows end-to-end learning of image/text encoders is multi-layer perceptron (MLP). The MLPs usually learn parametric embeddings to encode categorical/continuous data features. But due to their shallow architecture and context-free nature of the learned embeddings, they have the following limitations: (a) neither the model nor the learned embeddings are interpretable; (b) it is not robust against missing and noisy data (Section 3.2); (c) for semi-supervised learning, they do not achieve competitive performance (Section 3.4). Most importantly, the prediction accuracy of MLPs do not match that of tree-based models on most of the datasets (Arik & Pfister, 2019). To bridge this performance gap, researchers

---
*Corresponding author

have proposed various deep learning models (Arik & Pfister, 2019; Song et al., 2019; Cheng et al., 2016; Guo et al., 2018). Although these deep learning models achieve comparable prediction accuracy, they do not address all the limitations of GBDT and MLP. Furthermore, their comparisons are done in a limited setting of a handful of datasets. In particular, in Section 3.3 we show that when compared to standard GBDT on a large collection of datasets, GBDT perform significantly better than these recent models.

Different from tabular domain, the application of embeddings has been studied extensively in natural language processing. The embedding technique encodes discrete words (a categorical variable) in a dense low dimensional space, beginning from Word2Vec (Rong, 2014) with the context-free word embeddings to BERT (Devlin et al., 2019) which provides the contextual word embeddings. Based on contextual embedding, the self-attention Transformers (Vaswani et al., 2017) has achieved state-of-the-art performance on many NLP tasks. Additionally, the pre-training/fine-tuning paradigm in BERT, which pre-trains the Transformers on a large coprus of unsupervised text and fine-tunes it on downstream tasks with labeled text, has shed light on tabular data modeling in semi-supervised learning.

Motivated by the successful applications of Transformers in NLP, we adapt them in tabular domain. Particularly, TabTransformer modifies a sequence of multi-head attention-based Transformer layers on parametric embeddings to transform them into contextual embeddings, bridging the performance gap between baseline MLP and GBDT models. We investigate the effectiveness and interpretability of the resulting contextual embeddings generated by the Transformers. We find that highly correlated features (including feature pairs in the same column and cross column) result in embedding vectors that are close together in Euclidean distance, whereas no such pattern exists in context-free embeddings learned in a baseline MLP model. We also study the robustness of the TabTransformer against random missing and noisy data. The contextual embeddings make them highly robust in comparison to MLPs. Finally, we exploit the pre-training/fine-tuning methodologies from NLP and propose a semi-supervised learning approach for pre-training TabTransformer using unlabeled data and fine-tuning it on labeled data.

One of the key benefits of our proposed method for semi-supervised learning is the two independent training phases: a costly pre-training phase on unlabeled data and a lightweight fine-tuning phase on labeled data. This differs from many state-of-the-art semi-supervised methods (Chapelle et al., 2009; Oliver et al., 2018; Stretcu et al., 2019) that require a single training job including both the labeled and unlabeled data. The separated training procedure benefits the scenario where the model needs to be pretrained once but fine-tuned multiple times for multiple target variables. This scenario is in fact quite common in the industrial setting as companies tend to have one large dataset (e.g. describing customers/products) and are interested in applying multiple analyses on this data. We summarize our contributions as follows:

1. We propose TabTransformer, an architecture that provides and exploits contextual embeddings of categorical features. We provide extensive experiments showing that TabTransformer is superior to SoTA deep network models.
2. We investigate the resulting contextual embeddings and highlight their interpretability, contrasted to parametric context-free embeddings achieved by existing art.
3. We demonstrate the robustness of TabTransformer against noisy and missing data.
4. We provide and extensively study a two-phase pre-training then fine-tune procedure for tabular data, beating the state-of-the-art performance of semi-supervised learning methods.

## 2 Architecture and Training Process

The TabTransformer architecture comprises a column embedding layer, a stack of $N$ Transformer layers, and a multi-layer perceptron. Each Transformer layer (Vaswani et al., 2017) consists of a multi-head self-attention layer followed by a position-wise feed-forward layer. The architecture of TabTransformer is shown below in Figure 1.

In our experiments we use standard feature engineering techniques to transform special types such as text, zipcodes, ip addresses etc., into either numeric or categorical features. Although better techniques may exist for handling special data types they are outside the scope of this paper.

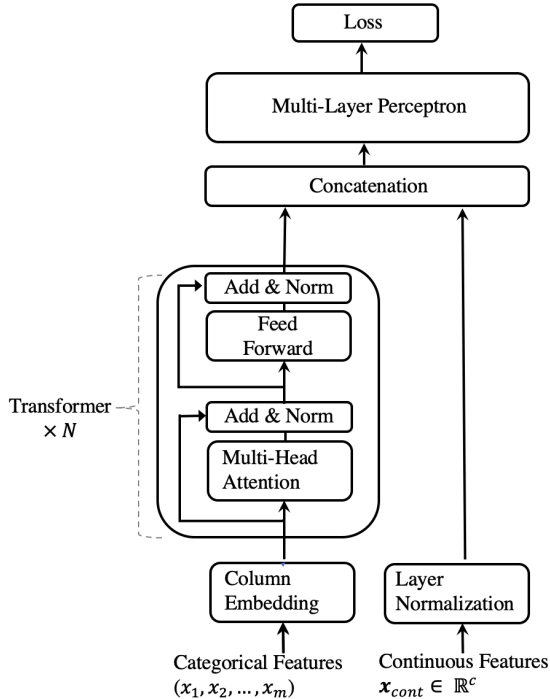

Figure 1: The architecture of TabTransformer.

Let $(\boldsymbol{x}, y)$ denote a features-target pair, where $\boldsymbol{x} \equiv \{\boldsymbol{x}_{\text{cat}}, \boldsymbol{x}_{\text{cont}}\}$ are processed features, and $y$ is target value. The $\boldsymbol{x}_{\text{cat}}$ denotes all the categorical features and $\boldsymbol{x}_{\text{cont}} \in \mathbb{R}^c$ denotes all of the $c$ continuous features. Let $\boldsymbol{x}_{\text{cat}} \equiv \{x_1, x_2, \cdots, x_m\}$ with each $x_i$ being a categorical feature, for $i \in \{1, \cdots, m\}$. We embed each of the $x_i$ categorical features into a parametric embedding of dimension $d$ using *Column embedding*, which is explained below in detail. Let $\boldsymbol{e}_{\phi_i}(x_i) \in \mathbb{R}^d$ for $i \in \{1, \cdots, m\}$ be the embedding of the $x_i$ feature, and $\boldsymbol{E}_\phi(\boldsymbol{x}_{\text{cat}}) = \{\boldsymbol{e}_{\phi_1}(x_1), \cdots, \boldsymbol{e}_{\phi_m}(x_m)\}$ be the set of embeddings for all the categorical features.

Next, these parametric embeddings $\boldsymbol{E}_\phi(\boldsymbol{x}_{\text{cat}})$ are passed through $N$ Transformer layers. Each parametric embedding is transformed into contextual embedding when outputted from the top layer Transformer, through successive aggregation of context from other embeddings. We denote the sequence of $N$ Transformer layers as a function $f_\theta$. The function $f_\theta$ operates on parametric embeddings $\{\boldsymbol{e}_{\phi_1}(x_1), \cdots, \boldsymbol{e}_{\phi_m}(x_m)\}$ and returns the corresponding contextual embeddings $\{\boldsymbol{h}_1, \cdots, \boldsymbol{h}_m\}$ where $\boldsymbol{h}_i \in \mathbb{R}^d$ for $i \in \{1, \cdots, m\}$. The contextual embeddings $\{\boldsymbol{h}_1, \cdots, \boldsymbol{h}_m\}$ are concatenated along with the continuous features $\boldsymbol{x}_{\text{cont}}$ to form a vector of dimension $(d \times m + c)$. This vector is inputted to an MLP, denoted by $g_\psi$, to predict the target $y$. Let $H$ be the cross-entropy for classification tasks and mean square error for regression tasks. We minimize the following loss function $\mathcal{L}(\boldsymbol{x}, y)$ to learn all the TabTransformer parameters in an end-to-end learning by the first-order gradient methods. The TabTransformer parameters include $\phi$ for column embedding, $\theta$ for Transformer layers, and $\psi$ for the top MLP layer.

$$\mathcal{L}(\boldsymbol{x}, y) \equiv H(g_{\boldsymbol{\psi}}(f_{\boldsymbol{\theta}}(\boldsymbol{E}_\phi(\boldsymbol{x}_{\text{cat}})), \boldsymbol{x}_{\text{cont}}), y). \quad (1)$$

Below, we explain the column embedding.

**Column embedding.** For each categorical feature (column) $i$, we have an embedding lookup table $\boldsymbol{e}_{\phi_i}(.)$, for $i \in \{1, 2, ..., m\}$. For $i$th feature with $d_i$ classes, the embedding table $\boldsymbol{e}_{\phi_i}(.)$ has $(d_i + 1)$ embeddings where the additional embedding corresponds to a missing value. The embedding for the encoded value $x_i = j \in [0, 1, 2, .., d_i]$ is $\boldsymbol{e}_{\phi_i}(j) = [\boldsymbol{c}_{\phi_i}, \boldsymbol{w}_{\phi_{ij}}]$, where $\boldsymbol{c}_{\phi_i} \in \mathbb{R}^\ell$ and $\boldsymbol{w}_{\phi_{ij}} \in \mathbb{R}^{d-\ell}$. The column-specific and unique identifier $\boldsymbol{c}_{\phi_i} \in \mathbb{R}^\ell$ distinguishes the classes in column $i$ from those in the other columns. The dimension of $\boldsymbol{c}_{\phi_i}$, $\ell$, is a hyper-parameter.

3

The use of unique identifier is innovative and particularly designed for tabular data. Rather in NLP, embeddings are element-wisely added with the positional encoding of the word in the sentence. Since, in tabular data, there is no ordering of the features, we do not use positional encodings. The strategies include both different choices for $\ell, d$ and element-wise adding the unique identifier and feature-value specific embeddings rather than concatenating them.

**Pre-training the Embeddings.** The contextual embeddings explained above are learned in end-to-end supervised training using labeled examples. For a scenario, when there are a few labeled examples and a large number of unlabeled ones, we introduce a pre-training procedure to train the Transformer layers using unlabeled data. This is followed by fine-tuning of the pre-trained Transformer layers along with the top MLP layer using the labeled data. For fine-tuning, we use the supervised loss defined in Equation 1.

We explore two different types of pre-training procedures, the masked language modeling (MLM) (Devlin et al., 2019) and the replaced token detection (RTD) (Clark et al., 2020). Given an input $\boldsymbol{x}_{\text{cat}} = \{x_1, x_2, ..., x_m\}$, MLM randomly selects $k\%$ features from index 1 to $m$ and masks them as missing. The Transformer layers along with the column embeddings are trained by minimizing cross-entropy loss of a multi-class classifier, which predicts the original features of the masked features using contextual embeddings outputted from the top-layer Transformer.

Instead of masking features, RTD replaces the original feature by a random value of that feature. Here, the loss is minimized for a binary classifier that predicts whether or not the feature has been replaced. To compute the replacement value, the original RTD in Clark et al. (2020) uses an encoder network to sample a subset of features. The reason to use an encoder network is that there are tens of thousands of tokens in language data and a uniformly random token can be easily detected. In contrast, we use uniformly random values to replace tabular features because (a) the number of classes within each categorical feature is typically limited; (b) a different binary classifier is defined for each column rather than a shared one, as each column has its own embedding lookup table. We name the two pre-training methods as TabTransformer-MLM and TabTransformer-RTD. In our experiments, the replacement value $k$ is set to 30.

## 3 EXPERIMENTS

**Data.** We evaluate TabTransformer and baseline models on 15 publicly available binary classification datasets from the UCI repository (Dua & Graff, 2017), the AutoML Challenge (Guyon et al., 2019), and Kaggle (Kaggle, 2020) for both supervised and semi-supervised learning. Each dataset is divided into five cross-validation splits. The training/validation/testing proportion of the data for each split are $65/15/20\%$. The number of categorical features across dataset ranges from 2 to 136. In the semi-supervised experiments, for each dataset and split, $p$ observations in the training data are uniformly sampled as the labeled data with a fixed random seed and the remaining training data are set as unlabeled set. The value of $p$ is chosen as 50, 200, and 500, corresponding to 3 different scenarios. In the supervised experiments, each training dataset is fully labeled.

**Setup.** For TabTransformer, the hidden (embedding) dimension, the number of layers and the number of attention heads are fixed to 32, 6, and 8 respectively; these parameters are pre-selected by hyperparameter optimization (HPO) on a small number of datasets. The MLP layer sizes are set to $\{4 \times l, 2 \times l\}$, where $l$ is the size of its input. Each competitor model is given 20 HPO rounds for each cross-validation split. For evaluation metrics, we use the Area under the curve (AUC) (Bradley, 1997). Note, the pre-training is only applied in semi-supervised scenario. We do not find much benefit in using it when the entire data is labeled. Its benefit is evident when there is a large number of unlabeled examples and a few labeled examples. Since in this scenario the pre-training provides a representation of the data that could not have been learned based only on the labeled examples.

### 3.1 THE EFFECTIVENESS OF THE TRANSFORMER LAYERS

First, a comparison between TabTransformers and the baseline MLP is conducted in a supervised learning scenario. We remove the Transformer layers $f_{\boldsymbol{\theta}}$ from the architecture, fix the rest of the components, and compare it with the original TabTransformer. The model without the Transformer layers is equivalently an MLP. The dimension of embeddings $d$ for categorical features is set as 32 for both models. The comparison results over 15 datasets are presented in Table 1. The TabTransformer

Table 1: Comparison between TabTransfomers and the baseline MLP. The evaluation metric is AUC in percentage.

| Dataset | Baseline MLP | TabTransformer | Gain (%) |
| --- | --- | --- | --- |
| albert | 74.0 | 75.7 | **1.7** |
| 1995_income | 90.5 | 90.6 | **0.1** |
| dota2games | 63.1 | 63.3 | **0.2** |
| hcdr_main | 74.3 | 75.1 | **0.8** |
| adult | 72.5 | 73.7 | **1.2** |
| bank_marketing | 92.9 | 93.4 | **0.5** |
| blastchar | 83.9 | 83.5 | **-0.4** |
| insurance_co | 69.7 | 74.4 | **4.7** |
| jasmine | 85.1 | 85.3 | **0.2** |
| online_shoppers | 91.9 | 92.7 | **0.8** |
| philippine | 82.1 | 83.4 | **1.3** |
| qsar_bio | 91.0 | 91.8 | **0.8** |
| seismicbumps | 73.5 | 75.1 | **1.6** |
| shrutime | 84.6 | 85.6 | **1.0** |
| spambase | 98.4 | 98.5 | **0.1** |

with the Transformer layers outperforms the baseline MLP on 14 out of 15 datasets with an average 1.0% gain in AUC.

Next, we take contextual embeddings from different layers of the Transformer and compute a t-SNE plot (Maaten & Hinton, 2008) to visualize their similarity in function space. More precisely, for each dataset we take its test data, pass their categorical features into a trained TabTransformer, and extract all contextual embeddings (across all columns) from a certain layer of the Transformer. The t-SNE algorithm is then used to reduce each embedding to a 2D point in the t-SNE plot. Figure 2 (left) shows the 2D visualization of embeddings from the last layer of the Transformer for dataset *bank_marketing*. Each marker in the plot represents an average of 2D points over the test data points for a certain class. We can see that semantically similar classes are close with each other and form clusters (annotated by a set of labels) in the embedding space. For example, all of the client-based features (color markers) such as job, education level and martial status stay close in the center and non-client based features (gray markers) such as month (last contact month of the year), day (last contact day of the week) lie outside the central area; in the bottom cluster the embedding of owning a housing loan stays close with that of being default; over the left cluster, embeddings of being a student, martial status as single, not having a housing loan, and education level as tertiary get together; and in the right cluster, education levels are closely associated with the occupation types (Torpey & Watson, 2014). In Figure 2, the center and right plots are t-SNE plots of embeddings before being passed through the Transformer and the context-free embeddings from MLP, respectively. For the embeddings before being passed into the Transformer, it starts to distinguish the non-client based features (gray markers) from the client-based features (color markers). For the embeddings from MLP, we do not observe such pattern and many categorical features which are not semantically similar are grouped together, as indicated by the annotation in the plot. We also evaluate the effectiveness of TabTransformer by fitting embeddings extracted from different layers into a linear model.

### 3.2 THE ROBUSTNESS OF TABTRANSFORMER

We demonstrate the robustness of TabTransformer on the noisy data and data with missing values, against the baseline MLP. We consider these two scenarios only on categorical features to specifically prove the robustness of contextual embeddings from the Transformer layers.

**Noisy Data.** On the test examples, we firstly contaminate them by replacing a certain number of values by randomly generated ones from the corresponding columns (features). Next, the noisy data are passed into a trained TabTransformer to compute a prediction AUC score. Results on a set of 3 datasets are presented in Figure 3. As the noisy rate increases, TabTransformer performs better

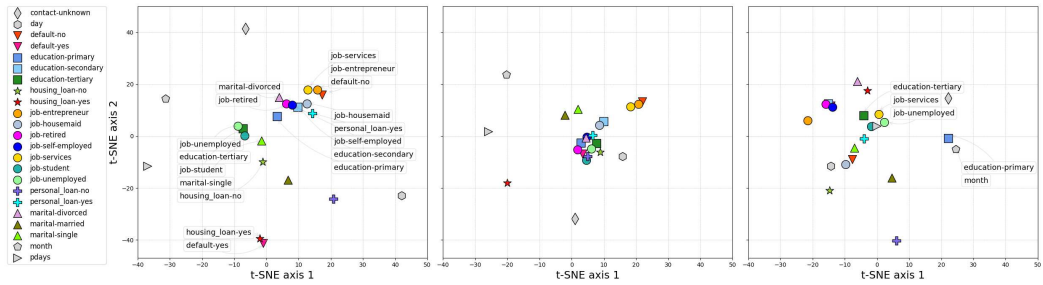

Figure 2: t-SNE plots of learned embeddings for categorical features on dataset *BankMarketing*. **Left**: TabTransformer – the embeddings generated from the last layer of the attention-based Transformer. **Center**: TabTransformer – the embeddings before being passed into the attention-based Transformer. **Right**: The embeddings learned from MLP.

in prediction accuracy and thus is more robust than MLP. In particular notice the *Blastchar* dataset where the performance is near identical with no noise, yet as the noise increases, TabTransformer becomes significantly more performant compared to the baseline. We conjecture that the robustness comes from the contextual property of the embeddings. Despite a feature being noisy, it draws information from the correct features allowing for a certain amount of correction.

**Data with Missing Values.** Similarly, on the test data we artificially select a number of values to be missing and send the data with missing values to a trained TabTransformer to compute the prediction score. Figure 4 shows the same patterns of the noisy data case, i.e. that TabTransformer shows better stability than MLP in handling missing values.

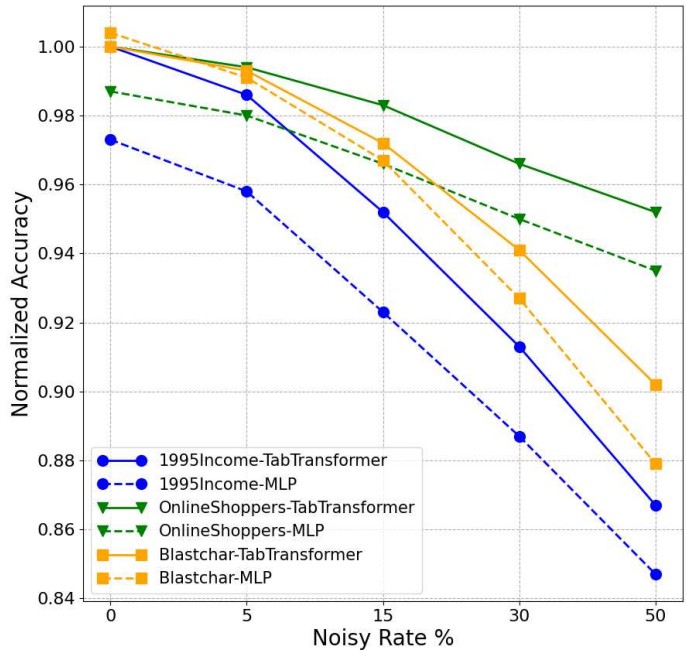

Figure 3: Performance of TabTransformer and MLP with noisy data. For each dataset, each prediction score is normalized by the score of TabTransformer at $0$ noise.

6

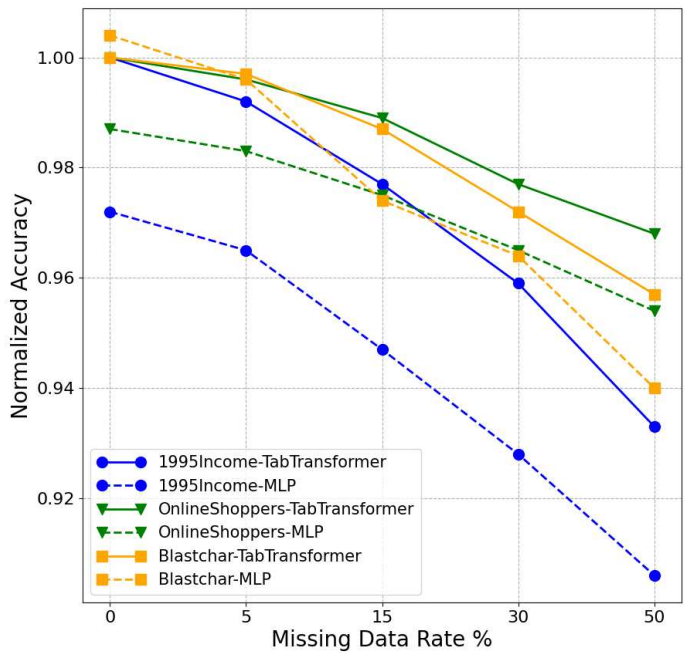

Figure 4: Performance of TabTransformer and MLP under missing data scenario. For each dataset, each prediction score is normalized by the score of TabTransformer trained without missing values.

Table 2: Model performance in supervised learning. The evaluation metric is mean $\pm$ standard deviation of AUC score over the 15 datasets for each model. Larger the number, better the result. The top 2 numbers are bold.

| Model Name | Mean AUC (%) |
| --- | --- |
| TabTransformer | **82.8** $\pm$ 0.4 |
| MLP | 81.8 $\pm$ 0.4 |
| GBDT | **82.9** $\pm$ 0.4 |
| Sparse MLP | 81.4 $\pm$ 0.4 |
| Logistic Regression | 80.4 $\pm$ 0.4 |
| TabNet | 77.1 $\pm$ 0.5 |
| VIB | 80.5 $\pm$ 0.4 |

### 3.3 SUPERVISED LEARNING

Here we compare the performance of TabTransformer against following four categories of methods: (a) Logistic regression and GBDT; (b) MLP and a sparse MLP following Morcos et al. (2019); (c) TabNet model of Arik & Pfister (2019); and (d) the Variational Information Bottleneck model (VIB) of Alemi et al. (2017).

Results are summarized in Table 2. TabTransformer, MLP, and GBDT are the top 3 performers. The TabTransformer outperforms the baseline MLP with an average 1.0% gain and perform comparable with the GBDT. Furthermore, the TabTransformer is significantly better than TabNet and VIB, the recent deep networks for tabular data.

### 3.4 SEMI-SUPERVISED LEARNING

We evaluate the TabTransformer under the semi-supervised learning scenario where few labeled training examples are available together with a significant number of unlabeled samples. Specifically, we compare our pretrained and then fine-tuned TabTransformer-RTD/MLM against following

Table 3: Semi-supervised learning results for 6 datasets with more than 30K data points, for different number of labeled data points. Evaluation metrics are mean AUC in percentage. Larger the number, better the result.

| # Labeled data | 50 | 200 | 500 |
| --- | --- | --- | --- |
| TabTransformer-RTD | $66.6 \pm 0.6$ | $70.9 \pm 0.6$ | $\mathbf{73.1} \pm 0.6$ |
| TabTransformer-MLM | $\mathbf{66.8} \pm 0.6$ | $\mathbf{71.0} \pm 0.6$ | $72.9 \pm 0.6$ |
| MLP (ER) | $65.6 \pm 0.6$ | $69.0 \pm 0.6$ | $71.0 \pm 0.6$ |
| MLP (PL) | $65.4 \pm 0.6$ | $68.8 \pm 0.6$ | $71.0 \pm 0.6$ |
| TabTransformer (ER) | $62.7 \pm 0.6$ | $67.1 \pm 0.6$ | $69.3 \pm 0.6$ |
| TabTransformer (PL) | $63.6 \pm 0.6$ | $67.3 \pm 0.7$ | $69.3 \pm 0.6$ |
| MLP (DAE) | $65.2 \pm 0.5$ | $68.5 \pm 0.6$ | $71.0 \pm 0.6$ |
| GBDT (PL) | $56.5 \pm 0.5$ | $63.1 \pm 0.6$ | $66.5 \pm 0.7$ |

semi-supervised models: (a) Entropy Regularization (ER) (Grandvalet & Bengio, 2006) combined with MLP and TabTransformer; (b) Pseudo Labeling (PL) (Lee, 2013) combined with MLP, Tab-Transformer, and GBDT (Jain, 2017); (c) MLP (DAE): an unsupervised pre-training method designed for deep models on tabular data – the swap noise Denoising AutoEncoder (Jahrer, 2018).

The pre-training models TabTransformer-MLM, TabTransformer-RTD and MLP (DAE) are firstly pretrained on the entire unlabeled training data and then fine-tuned on labeled data. The semi-supervised learning methods, Pseudo Labeling and Entropy Regularization, are trained on the mix of labeled and unlabeled training data. To better present results, we split the set of 15 datasets into two subsets. The first set includes 6 datasets with more than 30K data points and the second set includes remaining 9 datasets.

The results are presented in Table 3 and Table 4. When the number of unlabeled data is large, Table 3 shows that our TabTransformer-RTD and TabTransformer-MLM significantly outperform all the other competitors. Particularly, TabTransformer-RTD/MLM improves over all the other competitors by at least 1.2%, 2.0% and 2.1% on mean AUC for the scenario of 50, 200, and 500 labeled data points respectively. The Transformer-based semi-supervised learning methods TabTransformer (ER), TabTransformer (PL), and the tree-based semi-supervised learning method GBDT (PL) perform worse than the average of all the models. When the number of unlabeled data becomes smaller, as shown in Table 4, TabTransformer-RTD still outperforms most of its competitors but with a marginal improvement.

Furthermore, we observe that when the number of unlabeled data is small as shown in Table 4, TabTransformer-RTD performs better than TabTransformer-MLM, thanks to its easier pre-training task (a binary classification) than that of MLM (a multi-class classification). This aligns with the finding of the ELECTRA paper (Clark et al., 2020). In Table 4, with only 50 labeled data points, MLP (ER) and MLP (PL) beat our TabTransformer-RTD/MLM. This can be attributed to that there is room to improve in our fine-tuning procedure. Particularly, our approach allows to obtain informative embeddings but does not allow the weights of the classifier itself to be trained with unlabelled data. Since this issue does not occur for ER and PL, they obtain an advantage in extremely small labelled set. We point out however that this only means that the methods are complementary and a possible follow up could combine the best of all approaches.

Both evaluation results, Table 3 and Table 4, show that our TabTransformer-RTD and Transformers-MLM models are promising in extracting useful information from unlabeled data to help supervised training, and are particularly useful when the size of unlabeled data is large.

## 4 RELATED WORK

For supervised learning, standard MLPs have been applied to tabular data for many years (De Brébisson et al., 2015). For deep models designed specifically for tabular data, there are deep versions of factorization machines (Guo et al., 2018; Xiao et al., 2017), deep MLPs-based methods (Wang et al., 2017; Cheng et al., 2016; Cortes et al., 2016), Transformers-based methods (Song et al., 2019; Li et al., 2020; Sun et al., 2019), and deep versions of decision-tree-based algorithms

Table 4: Semi-supervised learning results for 9 datasets with less than 30K data points, for different number of labeled data points. Evaluation metrics are mean AUC in percentage. Larger the number, better the result.

| # Labeled data | 50 | 200 | 500 |
| --- | --- | --- | --- |
| TabTransformer-RTD | $78.6 \pm 0.6$ | $\mathbf{81.6} \pm 0.5$ | $\mathbf{83.4} \pm 0.5$ |
| TabTransformer-MLM | $78.5 \pm 0.6$ | $81.0 \pm 0.6$ | $82.4 \pm 0.5$ |
| MLP (ER) | $\mathbf{79.4} \pm 0.6$ | $81.1 \pm 0.6$ | $82.3 \pm 0.6$ |
| MLP (PL) | $79.1 \pm 0.6$ | $81.1 \pm 0.6$ | $82.0 \pm 0.6$ |
| TabTransformer (ER) | $77.9 \pm 0.6$ | $81.2 \pm 0.6$ | $82.1 \pm 0.6$ |
| TabTransformer (PL) | $77.8 \pm 0.6$ | $81.0 \pm 0.6$ | $82.1 \pm 0.6$ |
| MLP (DAE) | $78.5 \pm 0.7$ | $80.7 \pm 0.6$ | $82.2 \pm 0.6$ |
| GBDT (PL) | $73.4 \pm 0.7$ | $78.8 \pm 0.6$ | $81.3 \pm 0.6$ |

(Ke et al., 2019; Yang et al., 2018). In particular, Song et al. (2019) apply one layer of multi-head attention on embeddings to learn higher order features. The higher order features are concatenated and inputted to a fully connected layer to make the final prediction. Li et al. (2020) use self-attention layers and track the attention scores to obtain feature importance scores. Sun et al. (2019) combine the Factorization Machine model with transformer mechanism. All 3 papers are focused on recommendation systems with input data being high dimensional and extremely sparse, which makes it hard to have a clear comparison with this paper. Recent TabNet (Arik & Pfister, 2019) is designed on the sparse feature interaction of tabular data, and has a very different mechanism than our self-attention based one. There are a few works focusing on database oriented task in tabular domain, such as column categorization (Chen et al., 2019), entity linking (Luo et al., 2018), table layout identification (Habibi et al., 2020), and table augmentation (Deng et al., 2019). However, these tasks are fundamentally different from typical ML classification problem. They do not classify individual rows of a table, but properties of the table itself, i.e., meta-data. For this reason we do not elaborate the details of these papers nor compare our results with theirs.

For semi-supervised learning, Izmailov et al. (2019) give a semi-supervised method based on density estimation and evaluate their approach on tabular data. *Pseudo labeling* (Lee, 2013) is an efficient and popular baseline method. The pseudo labeling uses the current network to infer pseudo-labels of unlabeled examples, by choosing the most confident class. These pseudo-labels are treated like human-provided labels in the cross entropy loss. *Label propagation* (Zhu & Ghahramani, 2002; Iscen et al., 2019) is a similar approach where a node's labels propagate to all nodes according to their proximity, and are used by the training model as if they were the true labels. Another standard method in semi-supervised learning is *entropy regularization* (Grandvalet & Bengio, 2005; Sajjadi et al., 2016). It adds average per-sample entropy for the unlabeled examples to the original loss function for the labeled examples. Additionally, a classical approach of semi-supervised learning is co-training (Nigam & Ghani, 2000). However, the recent approaches - entropy regularization and pseudo labeling - are typically better and more popular. A succinct review of semi-supervised learning methods in general can be found in Oliver et al. (2019); Chappelle et al. (2010).

## 5 CONCLUSION

We proposed TabTransformer, a novel deep tabular data modeling architecture for supervised and semi-supervised learning. Extensive experiments show that TabTransformer significantly outperforms recent deep networks while matching the performance of GBDT. In addition, we study a two-phase pre-training/fine-tune paradigm for tabular data, beating the state-of-the-art semi-supervised learning methods. For future work, it would be interesting to investigate them in detail.

# DEPENDENCY STRUCTURE MISSPECIFICATION IN MULTI-SOURCE WEAK SUPERVISION MODELS

**Salva Rühling Cachay**
Technical University of Darmstadt
salvaruehling@gmail.com

**Benedikt Boecking**
Carnegie Mellon University

**Artur Dubrawski**
Carnegie Mellon University

## ABSTRACT

Data programming (DP) has proven to be an attractive alternative to costly hand-labeling of data. In DP, users encode domain knowledge into *labeling functions* (LF), heuristics that label a subset of the data noisily and may have complex dependencies. A label model is then fit to the LFs to produce an estimate of the unknown class label. The effects of label model misspecification on test set performance of a downstream classifier are understudied. This presents a serious awareness gap to practitioners, in particular since the dependency structure among LFs is frequently ignored in field applications of DP. We analyse modeling errors due to structure over-specification. We derive novel theoretical bounds on the modeling error and empirically show that this error can be substantial, even when modeling a seemingly sensible structure.

## 1 INTRODUCTION

Annotating large datasets for machine learning is expensive, time consuming, and a bottleneck for many practical applications of artificial intelligence. Recently, data programming, a paradigm that makes use of multiple weak supervision sources, has emerged as a promising alternative to manual data annotation (Ratner et al., 2016). In this framework, users encode domain knowledge into weak supervision sources, such as domain heuristics or knowledge bases, that each noisily label a subset of the data. A generative model over the sources and the latent true label is then learned. One can use the learned model to estimate *probabilistic* labels to train a *downstream model*, replacing the need to obtain ground truth labels by manual annotation of individual samples.

In practice, the sources of weak labels that users define often exhibit statistical dependencies amongst each other, e.g. sources operating on the same or similar input (some examples can be found in Table 1). Defining the correct dependency structure is difficult, thus a common approach in popular libraries (Bach et al., 2019; Ratner et al., 2019a) and related research (Dawid & Skene, 1979; Anandkumar et al., 2014; Varma & Ré, 2018; Boecking et al., 2021) is to ignore it. However, the implications of this assumption on downstream performance have not been researched in detail. Therefore, in this paper we take steps towards gaining a better understanding of the trade-offs involved.

**Contributions** We present novel bounds on the label model posterior and the downstream generalization risk that are explicitly influenced by misspecified higher-order dependencies. We also introduce three new higher-order dependency types, which we name *bolstering*, *negated*, and *priority* dependencies. Lastly, we empirically show that downstream test performance is highly sensitive to the user-specified dependencies, even when they make sense semantically. The finding suggests that in practice, it is advisable to only carefully model a few, if any, dependencies.

## 2 RELATED WORK

**Data programming** While the original data programming framework (Ratner et al., 2016) is based on a factor graph that support the modeling of arbitrary dependencies between LFs, more recent methods for solving for the parameters of the label model only support the modeling of pairwise correlations (Ratner et al., 2019a;b; Fu et al., 2020) — as such, losing some of the expressive power of data programming. The former extends data programming to the multi-task setting by exploiting the graph structure of the inverse covariance matrix among the sources (Ratner et al., 2019b) — in particular the fact that an entry is zero when there is no edge between the corresponding sources in the graphical model (Loh & Wainwright, 2012). The latter finds a closed-form solution for a class of

binary Ising models by using triplet methods (Fu et al., 2020). Our experimental findings suggest that practitioners may indeed benefit from simply ignoring higher-order dependencies.

**Structure learning** In order to automatically learn the structure between these sources, previous work optimizes the marginal pseudolikelihood of the noisy labels (Bach et al., 2017), or makes use of robust PCA to denoise the inverse covariance matrix of the sources labels into a graph structured term (Varma et al., 2019). A different approach, infers the structure through static analysis of the weak supervision sources code definitions and thereby reduces the sample complexity for learning the structure (Varma et al., 2017). In our experiments, we show that such methods should be carefully used, and may in fact lead to downstream performance losses.

**Model misspecification** On the side of work on model misspecification, (White, 1982) establishes that the Maximum Likelihood Estimator of a misspecified model is a consistent estimator of the learnable parameter that minimizes the KL divergence to the true distribution – if that optimal, misspecified parameter is globally identifiable. In an interesting result, (Jog & Loh, 2015) show that the KL divergence between a multivariate Gaussian distribution and a misspecified (by at least a single edge) Gaussian graphical model is bounded by a constant from below. It emphasizes the need to correctly select the model's edge structure, since otherwise the fitted distribution will differ from the true one in terms of KL divergence.

## 3 Problem setup

For this work we use the data programming framework as introduced in (Ratner et al., 2016), where the premise is that experts can model any higher-order dependency between labeling functions. We extend it by *negated*, *bolstering*, *priority* dependencies (definitions in the appendix B), e.g. the latter encoding the notion that one source's vote should be prioritized over the one from a noisier source. The additional dependency types are motivated by our selected LF sets that contain dependencies, such as the ones in Table 1, that we could not express before. Newer weak supervision models and model fitting approaches often only allow for pairwise correlation dependencies to be modeled (Ratner et al., 2019b; Fu et al., 2020).

Let $(x, y) \sim \mathcal{D}$ be the true data generating distribution and for simplicity assume that $y \in \{-1, 1\}$. As in (Ratner et al., 2016), users provide $m$ labeling functions (LFs) $\lambda = \lambda(x) \in \{-1, 0, 1\}^m$, where $0$ means that the LF abstained from labeling. Following (Ratner et al., 2016), we model the joint distribution of $y, \lambda$ as a factor graph. To study model misspecification we compare two label models, $p_\theta$ for the conditional independent case and $p_\mu$ which models $M$ higher-order dependencies:

$$p_\mu(\lambda, y) = \frac{1}{Z_\mu} \exp\left(\mu^T \phi(\lambda, y)\right) = Z_\mu^{-1} \exp\left(\mu_1^T \phi_1(\lambda, y) + \mu_2^T \phi_2(\lambda, y)\right), \qquad \mu \in \mathbb{R}^{m+M} \quad (1)$$

$$p_\theta(\lambda, y) = Z_\theta^{-1} \exp\left(\theta^T \phi_1(\lambda, y)\right), \qquad \theta \in \mathbb{R}^m, \quad (2)$$

where $\phi_1(\lambda, y) = \lambda y$ are the accuracy factors, $\phi_2(\cdot)$ are arbitrary, higher-order dependencies and $Z_\theta^{-1}, Z_\mu^{-1}$ are normalization constants. We assume w.l.o.g. that factors are bounded $\leq 1$. Note that for ease of exposition, the models above do not model the labeling propensity factor (also known as LF coverage). Since this factor does not depend on the label, the corresponding terms would cancel out in the quantities studied below (same goes for dependencies that do not depend on $y$, e.g. the *similar* factor from (Ratner et al., 2016)).

## 4 Theoretical analysis

**Bound on the label model posterior under model misspecification** We now state our bound on the probabilistic label difference (which we prove in the appendix A):

$$|p_\mu(y \mid \lambda) - p_\theta(y \mid \lambda)| \leq \frac{1}{2}||\mu_1 - \theta||_1 + \frac{1}{4}||\mu_2||_1 \quad (3)$$

This is an important quantity of interest since the probabilistic labels are used to train a downstream model. Unsurprisingly then, this quantity reappears as a main factor controlling the KL divergence and generalization risk, see below.

Suppose that the downstream model $f_w : \mathcal{X} \to \mathcal{Y}$ is parameterized by $w$, and that to learn $w$ we minimize a, w.l.o.g., bounded noise-aware loss function $L(f_w(x), y) \in [0, 1]$ (that acts on the probabilistic labels). If we had access to the true labels, we would normally try to find the $w$ that minimizes the risk: $w^* = \arg\min_w R(w) = \arg\min_w \mathbb{E}_{(x,y)\sim D}[L(f_w(x), y)]$. Since this is not the case, we instead will get the parameter $\hat{w}$ that minimizes the (empirical) noise-aware loss.

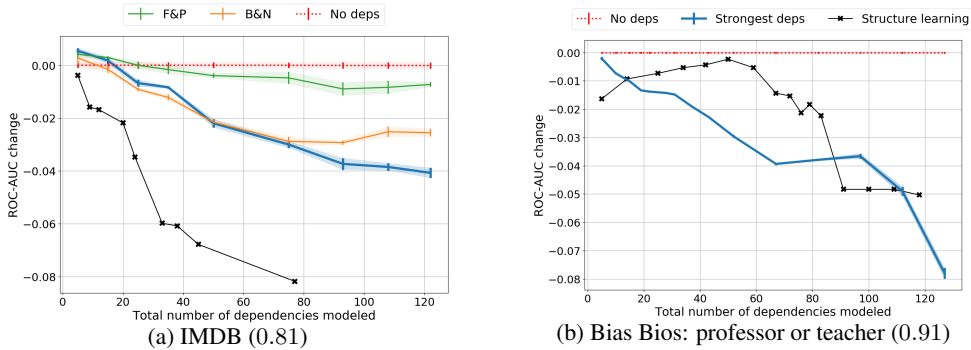

(a) IMDB (0.81)   (b) Bias Bios: professor or teacher (0.91)

Figure 1: Modeling more than a handful dependencies (as the ones in Table 1) significantly deteriorates ROC-AUC downstream performance as compared to simply ignoring them by up to 4 and 8 points. This effect intensifies as we model more dependencies. In brackets, the baseline score for the independent model ('No deps'). B&N means that we only model the *bolstering* and *negated* dependencies, while F&P means that we only model the *fixing* and *priority* dependencies. Structure learning means that we model the *similar*, *fixing* and *reinforcing* dependencies returned by the method from (Bach et al., 2017) for different threshold hyperparameters (the lower, the more dependencies are modeled).

**Bound on the generalization risk under model misspecification** While (Fu et al., 2020) provide a bound on the generalization error that accounts for model misspecification – the label model being a less expressive Ising model (i.e. no higher-order dependencies may be modeled) – the part of the bound corresponding to the misspecification is a somewhat loose KL-divergence term between the true and the misspecified models. If we instead assume that there exists a label model for some optimal parameterization of factors (not necessarily the ones that are actually modeled) that is equivalent to the true data generating distribution, we can provide a more meaningful and tight bound.

As in (Ratner et al., 2016; 2019b), assume that 1) there exists an optimal parameter of either $p_\mu$ or $p_\theta$ (say, $\mu^*$ with label model $p_{\mu^*}$) such that sampling $(\lambda, y)$ from this optimal label model is equivalent to $(\lambda, y) \sim \mathcal{D}$; and 2) the label $y$ is independent of the features used for training $f_w$ given the labeling function outputs $\lambda$. Differently than (Ratner et al., 2016; 2019b), the factors that are actually modeled may differ from the ones of the optimal label model. By adapting the proof of theorem 1 in (Ratner et al., 2019b) to our case where we incorrectly use a misspecified model (say, $p_\theta$), we can bound the generalization risk as follows

$$R(\hat{w}) - R(w^*) \leq \gamma + 2||\mu_1^* - \theta||_1 + ||\mu_2^*||_1, \quad (4)$$

where $\gamma$ is a bound on the empirical risk minimization error, which is not specific to our setting. We can bound the KL divergence of the two models by a similar quantity, see the appendix A.3.

**Interpretation** These bounds naturally involve the accuracy parameter estimation error $||\mu_1 - \theta||_1$ and the learned strength $s := ||\mu_2||_1$ of the dependencies only modeled in $p_\mu$. Note that if we assume that the model with dependencies $p_\mu$ is the true one we can interpret $s$ as the magnitude of the dependencies that the independent model $p_\theta$ failed to model. If on the other hand we associate the misspecified model with $p_\mu$, then we can interpret this quantity $s$ as the (dependency) parameter excess that was incorrectly learned by $p_\mu$. The presented bounds are tighter than the ones from (Ratner et al., 2019b) for the L1 norm, while in addition accounting for model misspecification.

## 5 EXPERIMENTS

### 5.1 PROXY FOR FINDING TRUE DEPENDENCIES IN REAL DATASETS

The underlying *true* structure between two real labeling functions $\lambda_j, \lambda_k$ is, of course, unknown. However, by using true training labels (solely for this purpose) together with the observed LF votes, we can compute resulting factor values for each data point $i$, to observe empirical strength of dependency factor $l$ over a training set: $v_{j,k}^l = \sum_i \phi^l(\lambda(x_i)_j, \lambda(x_i)_k, y_i)$. Sorting dependencies $l$

Table 1: The strongest and weakest dependencies among the IMDB LFs, as measured by their factor value $v_{j,k}^l$ computed with respect to the true labels. In the experiment from Fig.1, we repeatedly add weaker dependencies to the set of dependencies used for learning the label model, starting with the strongest ones below.

| $LF_j$ | $LF_k$ | factor type $l$ | factor value $v_{j,k}^l$ |
|---|---|---|---|
| best | great | bolstering | 801 |
| bad | don't waste | bolstering | 110 |
| original | bad | priority | 327 |
| recommend | terrible | priority | 53 |
| worth | not worth | fixing | 238 |
| great | nothing great | fixing | 15 |
| worth | not worth | negated | 219 |
| special | not special | negated | 8 |
| recommend | highly recommend | reinforcing | 226 |
| bad | absolutely horrible | reinforcing | 7 |

according to $v_{j,k}^l$ in descending order, we then choose to model the top $d$ dependencies. These are the dependencies for which the true labels provide the most evidence of being correct.

## 5.2 Performance deterioration due to structure over-specification

For the following experiment we use the IMDB Movie Review Sentiment dataset consisting of $n = 25k$ training and test samples each (Maas et al., 2011) and manually select a set of $m = 135$ sensible LFs that label on the presence of a single word or a pair of words (i.e. uni-/bi-gram LFs). In addition we use the Bias in Bios dataset (De-Arteaga et al., 2019), and focus on the binary classification task introduced in (Boecking et al., 2021) that aims at distinguishing the biographies of professors and teachers ($n = 12294, m = 85$). We deliberately choose unigram and bigram LFs so as to create dependencies we expect to help with downstream model performance, e.g. by adding negations (e.g. "not worth") of unigrams (e.g. "worth") that we expect to fix the less precise votes of the latter when both do not abstain.

We choose different $d \in \{1, 3, 5, \ldots, 40\}$ and then model the strongest $\leq d$ dependencies of each factor $l$ according to $v^l$. An example of the strongest and weakest dependencies for the IMDB dataset is shown in Table 1. For the Bias in Bios experiment, an example of a strong bolstering dependency is that the term 'PhD' appears in addition to the term 'university'. We report the test set performance of a simple 3-layer neural network trained on the probabilistic labels, averaged out over 100 runs. While for IMDB we observe a marginal boost ($< 0.005$) in performance when modeling the strongest $d = 1, 3$ dependencies of each factor (5, 15 in total), the main take-away is the following:

We find that modeling more than a handful of dependencies significantly *deteriorates* the downstream end classifier performance (by up to 8 ROC-AUC points) as compared to simply ignoring them (Fig. 1). The performance worsens as we increase $d$, i.e. as we model more, slightly weaker, dependencies. We reiterate that these additional dependencies still, semantically, make sense (as depicted in Table 1, where the weakest ones are modeled only for the case where the total number of dependencies = 122). Using the structure learning method from (Bach et al., 2017) to infer the dependency structure results in worse test performances too.

## 6 Discussion and Future Work

**Discussion** Even though this result and insight is highly relevant for practitioners, it has, to the best of our knowledge, not been explored in detail. It may come as a surprise that modeling seemingly sensible dependencies can significantly deteriorate the targeted downstream model performance. We hypothesize that this is due to the true model being close to the conditionally independent case and in our presented bounds, we indeed see that they become looser as more incorrect dependencies are modeled. In addition, more complex models often suffer of a higher sample complexity, as is also briefly noted in (Fu et al., 2020). This suggests that practitioners may 1) indeed be best served, at first, by simply ignoring (higher-order) dependencies; and 2) need to be careful when specifying

dependencies, either by hand or through structure learning algorithms, which emphasizes the need for a small ground-truth labeled validation set to compare the performance of different label models.

**Future work** Future work should give a theoretically precise answer to the reason for why performance deterioration is observed, conduct more extensive experiments to validate that this holds for a variety of datasets and labeling function sets, as well as better characterize the settings in which structure learning actually helps with downstream performance.

## A  Proofs of the theoretical analysis

### A.1  Problem setup recap

Let $(x, y) \sim \mathcal{D}$ be the true data generating distribution and for simplicity assume that $y \in \mathcal{Y} = \{-1, 1\}$. Users provide $m$ labeling functions (LFs) $\lambda = \lambda(x) \in \{-1, 0, 1\}^m$, where 0 means that the LF abstained from labeling. We compare two label models, $p_\theta$ for the conditional independent case, and $p_\mu$ which models higher-order dependencies:

$$p_\mu(\lambda, y) = \frac{1}{Z_\mu} \exp\left(\mu^T \phi(\lambda, y)\right) = Z_\mu^{-1} \exp\left(\mu_1^T \phi_1(\lambda, y) + \mu_2^T \phi_2(\lambda, y)\right), \qquad \mu \in \mathbb{R}^{m+M} \quad (5)$$

$$p_\theta(\lambda, y) = Z_\theta^{-1} \exp\left(\theta^T \phi_1(\lambda, y)\right), \qquad \theta \in \mathbb{R}^m, \quad (6)$$

where $\phi_1(\lambda, y) = \lambda y$ are the accuracy factors, $\phi_2(\cdot)$ are arbitrary, higher-order dependencies and $Z_\theta^{-1}, Z_\mu^{-1}$ are normalization constants. We assume w.l.o.g. that factors are bounded $\leq 1$. Using an unlabeled dataset $X = \{x_i\}_{i=1}^n$ of n data points $x_i \in \mathcal{X}$ to which we each apply the $m$ user provided labeling functions, we attain the label matrix $\Lambda \in \{-1, 0, 1\}^{n \times m}$. With $\Lambda$ we then train the label model to get a set of $n$ probabilistic labels with which we supervise the downstream model $f_w : \mathcal{X} \to \mathcal{Y}$.

**Lemma 1** (Sigmoid posterior). With $\sigma(x) = \frac{1}{1+\exp(-x)}$ being the sigmoid function, it holds that

$$p_\mu(y \mid \lambda) = \sigma\left(2\mu_1^T \phi_1(\lambda, y) + \mu_2^T \left(\phi_2(\lambda, y) - \phi_2(\lambda, -y)\right)\right) \quad (7)$$

$$p_\theta(y \mid \lambda) = \sigma\left(2\theta^T \phi_1(\lambda, y)\right). \quad (8)$$

**Proof**

$$\begin{aligned}
p_\mu(y \mid \lambda) &= \frac{p_\mu(\lambda, y)}{p_\mu(\lambda)} = \frac{p_\mu(\lambda, y)}{\sum_{\tilde{y} \in \mathcal{Y}} p_\mu(\lambda, \tilde{y})} \\
&= \frac{Z_\mu^{-1} \exp\left(\mu^T \phi(\lambda, y)\right)}{\sum_{\tilde{y} \in \mathcal{Y}} Z_\mu^{-1} \exp\left(\mu^T \phi(\lambda, \tilde{y})\right)} \\
&= \frac{\exp\left(\mu^T \phi(\lambda, y)\right)}{\sum_{\tilde{y} \in \mathcal{Y}} \exp\left(\mu^T \phi(\lambda, \tilde{y})\right)} \\
&= \frac{\exp\left(\mu^T \phi(\lambda, y)\right)}{\exp\left(\mu^T \phi(\lambda, y)\right) + \exp\left(\mu^T \phi(\lambda, -y)\right)} \\
&= \frac{1}{1 + \exp\left(\mu^T \left(\phi(\lambda, -y) - \phi(\lambda, y)\right)\right)} \\
&= \sigma\left(\mu^T \left(\phi(\lambda, y) - \phi(\lambda, -y)\right)\right) \\
&= \sigma\left(2\mu_1^T \phi_1(\lambda, y) + \mu_2^T \left(\phi_2(\lambda, y) - \phi_2(\lambda, -y)\right)\right),
\end{aligned}$$

where in the last step we used the fact that the accuracy factors are odd functions, i.e. $\phi_1(\lambda, -y) = -\lambda y = -\phi_1(\lambda, y)$. Eq. 8 follows by the same argumentation.

A.2 PROOF OF THE BOUND ON THE LABEL MODEL POSTERIOR

**Bound**  Our bound on the probabilistic label difference between the two models above is:

$$|p_\mu(y \mid \lambda) - p_\theta(y \mid \lambda)| \leq \frac{1}{2}||\mu_1 - \theta||_1 + \frac{1}{4}||\mu_2||_1 \tag{9}$$

**Proof**  Using Lemma 1 we have that

$$|p_\mu(y \mid \lambda) - p_\theta(y \mid \lambda)| = \left|\sigma\left(2\mu_1^T \phi_1(\lambda, y) + \mu_2^T (\phi_2(\lambda, y) - \phi_2(\lambda, -y))\right) - \sigma\left(2\theta^T \phi_1(\lambda, y)\right)\right|$$

By the mean value theorem it follows that for some c between the arguments of $\sigma$ above

$$= \sigma'(c) \left|\left(2\mu_1^T \phi_1(\lambda, y) + \mu_2^T (\phi_2(\lambda, y) - \phi_2(\lambda, -y))\right) - 2\theta^T \phi_1(\lambda, y)\right|$$
$$= \sigma'(c) \left|2(\mu_1 - \theta)^T \phi_1(\lambda, y) + \mu_2^T (\phi_2(\lambda, y) - \phi_2(\lambda, -y))\right|$$

Using the triangle inequality and the fact that $\max_x \sigma'(x) = \max_x \sigma(x)(1 - \sigma(x)) = \frac{1}{4}$, we can now bound this expression as follows

$$\leq \frac{1}{2}\left|(\mu_1 - \theta)^T \phi_1(\lambda, y)\right| + \frac{1}{4}\left|\mu_2^T (\phi_2(\lambda, y) - \phi_2(\lambda, -y))\right|$$

finally, since the defined higher-order dependencies are indicator functions $\neq 0$ for only one $y \in \mathcal{Y}$, and if $||q||_\infty \leq 1$ then $|x^T q| = |\sum_i x_i q_i| \leq \sum_i |x_i q_i| \leq \sum_i |x_i| = ||x||_1$, this reduces to

$$\leq \frac{1}{2}||\mu_1 - \theta||_1 + \frac{1}{4}||\mu_2||_1.$$

A.3 PROOF OF THE BOUND ON THE KL DIVERGENCE

**Bound**
$$KL\left(p_\mu(y \mid \lambda) \mid\mid p_\theta(y \mid \lambda)\right) \leq 2||\mu_1 - \theta||_1 + ||\mu_2||_1 \tag{10}$$

**Proof**  We adapt Theorem 7 of (Honorio, 2011) to give a bound on the KL divergence between the two label model posterior's. First note that with the same line of argumentation as in A.2 we have that

$$|\log p_\mu(y \mid \lambda) - \log p_\theta(y \mid \lambda)| = (\log \sigma)'(c) \left|2(\mu_1 - \theta)^T \phi_1(\lambda, y) + \mu_2^T (\phi_2(\lambda, y) - \phi_2(\lambda, -y))\right|$$
$$\leq 2||\mu_1 - \theta||_1 + ||\mu_2||_1,$$

where we use that the derivative of $\log \sigma(\cdot)$ is $(1 + \exp(x))^{-1} \in (0, 1)$, and is bounded by 1. Next

$$KL\left(p_\mu(y \mid \lambda) \mid\mid p_\theta(y \mid \lambda)\right) = \sum_{\lambda \in L} p_\mu(\lambda) \sum_{y \in \mathcal{Y}} p_\mu(y|\lambda) \cdot \log\left(\frac{p_\mu(y|\lambda)}{p_\theta(y|\lambda)}\right)$$
$$= \sum_{\lambda \in L} p_\mu(\lambda) \sum_{y \in \mathcal{Y}} \frac{p_\mu(\lambda, y)}{p_\mu(\lambda)} \cdot \log\left(\frac{p_\mu(y|\lambda)}{p_\theta(y|\lambda)}\right)$$
$$= \sum_{\lambda \in L} \sum_{y \in \mathcal{Y}} p_\mu(\lambda, y) \cdot (\log p_\mu(y|\lambda) - \log p_\theta(y|\lambda))$$
$$\leq \sum_{\lambda \in L} \sum_{y \in \mathcal{Y}} p_\mu(\lambda, y) \cdot |\log p_\mu(y|\lambda) - \log p_\theta(y|\lambda)|$$
$$\leq (2||\mu_1 - \theta||_1 + ||\mu_2||_1) \sum_{\lambda \in L} \sum_{y \in \mathcal{Y}} p_\mu(\lambda, y)$$
$$= 2||\mu_1 - \theta||_1 + ||\mu_2||_1$$

A.4 PROOF OF THE GENERALIZATION RISK BOUND

We now adapt Theorem 1 from (Ratner et al., 2019b) to the setting with model misspecification and assume like in (Ratner et al., 2016; 2019b) that

1. there exists an optimal parameter of either $p_\mu$ or $p_\theta$ such that sampling $(\lambda, y)$ from this optimal label model is equivalent to $(\lambda, y) \sim \mathcal{D}$.
2. the label $y$ is independent of the features used for training $f_w$ given the labeling function outputs $\lambda$, i.e. the LF labels provide sufficient signal to identify the label.

For 1. we now assume without loss of generality, that $p_{\mu^*}(\lambda, y) = p_\mathcal{D}(\lambda, y)$ for an optimal parameter $\mu^* \in \mathbb{R}^{m+M}$, and that we incorrectly use the misspecified label model $p_\theta$. For the reversed roles (i.e. $p_\theta$ is the true model and $p_\mu$ is misspecified), the following arguments are symmetric.

Suppose that the downstream model $f_w : \mathcal{X} \to \mathcal{Y}$ is parameterized by $w$, and that to learn $w$ we minimize a, w.l.o.g., bounded loss function $L(f_w(x), y) \in [0, 1]$. If we had access to the true labels, we would normally try to find the $w$ that minimizes the risk:

$$w^* = \arg\min_w R(w) = \arg\min_w \mathbb{E}_{(x,y)\sim D}\left[L(f_w(x), y)\right]. \tag{11}$$

Since this is not the case, we instead minimize the noise-aware loss:

$$\tilde{w} = \arg\min_w R_\theta(w) = \arg\min_w \mathbb{E}_{(x,y)\sim\mathcal{D}}\left[\mathbb{E}_{\tilde{y}\sim p_\theta(\cdot|\lambda)}\left[L(f_w(x), \tilde{y})\right]\right]. \tag{12}$$

In practice we will get the parameter $\hat{w}$ that minimizes the empirical noise-aware loss over the unlabeled dataset $X = \{x_1, ..., x_n\}$:

$$\hat{w} = \arg\min_w \hat{R}_\theta(w) = \arg\min_w \frac{1}{n}\sum_{i=1}^n \mathbb{E}_{\tilde{y}\sim p_\theta(\cdot|\lambda(x_i))}\left[L(f_w(x_i), \tilde{y})\right]. \tag{13}$$

Since the empirical risk minimization error is not specific to our setting and can be done with standard methods, we simply assume that the error $|R_\theta(\tilde{w}) - R_\theta(\hat{w})| \leq \gamma(n)$, where $\gamma(n)$ is a decreasing function of the unlabeled dataset size $n$.

**Bound** By adapting the proof of theorem 1 in (Ratner et al., 2019b) to our case with model misspecification involved, we can bound the generalization risk as follows

$$R(\hat{w}) - R(w^*) \leq \gamma(n) + 2\|\mu_1^* - \theta\|_1 + \|\mu_2^*\|_1. \tag{14}$$

Note that the bound from (Ratner et al., 2019b) is mistakenly too tight by a factor of 2 (due to the last step in the proof below that is partly overseen).

**Proof** First, using the law of total expectation, followed by our assumptions 2. and 1., in that order, we have that:

$$\begin{aligned}
R(w) &= \mathbb{E}_{(x',y')\sim D}\left[R(w)\right] \\
&= \mathbb{E}_{(x',y')\sim D}\left[\mathbb{E}_{(x,y)\sim D}\left[L(f_w(x'), y)|x = x'\right]\right] \\
&= \mathbb{E}_{(x',y')\sim D}\left[\mathbb{E}_{(x,y)\sim D}\left[L(f_w(x'), y)|\lambda(x) = \lambda(x')\right]\right] \\
&= \mathbb{E}_{(x',y')\sim D}\left[\mathbb{E}_{\tilde{y}\sim p_{\mu^*}(\cdot|\lambda)}\left[L(f_w(x'), \tilde{y})\right]\right] \\
&= R_{\mu^*}(w)
\end{aligned}$$

Using the result above that $R = R_{\mu^*}$ and adding and subtracting terms we have:

$$\begin{aligned}
R(\hat{w}) - R(w^*) &= R_{\mu^*}(\hat{w}) - R_{\mu^*}(w^*) \\
&= R_{\mu^*}(\hat{w}) + R_\theta(\hat{w}) - R_\theta(\hat{w}) + R_\theta(\tilde{w}) - R_\theta(\tilde{w}) - R_{\mu^*}(w^*)
\end{aligned}$$

since $\tilde{w}$ minimizes the noise-aware risk, i.e. $R_\theta(\tilde{w}) \leq R_\theta(w^*)$, we have that:

$$\begin{aligned}
&\leq R_{\mu^*}(\hat{w}) + R_\theta(\hat{w}) - R_\theta(\hat{w}) + R_\theta(w^*) - R_\theta(\tilde{w}) - R_{\mu^*}(w^*) \\
&\leq |R_\theta(\hat{w}) - R_\theta(\tilde{w})| + |R_{\mu^*}(\hat{w}) - R_\theta(\hat{w})| + |R_\theta(w^*) - R_{\mu^*}(w^*)| \\
&\leq \gamma(n) + 2\max_{w'}|R_{\mu^*}(w') - R_\theta(w')|
\end{aligned}$$

The main term $|R_{\mu^*}(w') - R_\theta(w')|$ we now have in the bound, is the difference between the true/expected noise-aware losses given the true label model parameter $\mu^*$ and the estimated parameter

$\theta$ for the misspecified model. We now bound this quantity:

$$|R_{\mu^*}(w') - R_\theta(w')| = \left| \mathbb{E}_{(x,y) \sim \mathcal{D}} \left[ \mathbb{E}_{\tilde{y} \sim p_{\mu^*}(\cdot|\lambda)} [L(f_w(x), \tilde{y})] - \mathbb{E}_{\tilde{y} \sim p_\theta(\cdot|\lambda)} [L(f_w(x), \tilde{y})] \right] \right|$$

$$= \left| \mathbb{E}_{(x,y) \sim \mathcal{D}} \left[ \sum_{y' \in \mathcal{Y}} L(f_w(x), y') \left( p_{\mu^*}(y'|\lambda) - p_\theta(y'|\lambda) \right) \right] \right|$$

$$\leq \sum_{y' \in \mathcal{Y}} \mathbb{E}_{(x,y) \sim \mathcal{D}} \left[ |L(f_w(x), y') \left( p_{\mu^*}(y'|\lambda) - p_\theta(y'|\lambda) \right)| \right]$$

$$\leq \sum_{y' \in \mathcal{Y}} \mathbb{E}_{(x,y) \sim \mathcal{D}} \left[ |p_{\mu^*}(y'|\lambda) - p_\theta(y'|\lambda)| \right]$$

$$\leq |\mathcal{Y}| \max_{y'} \mathbb{E}_{(x,y) \sim \mathcal{D}} \left[ |p_{\mu^*}(y'|\lambda) - p_\theta(y'|\lambda)| \right]$$

We can now use our bound from (3) and get that:

$$\leq 2 \left( \frac{1}{2} ||\mu_1^* - \theta||_1 + \frac{1}{4} ||\mu_2^*||_1 \right) = ||\mu_1^* - \theta||_1 + \frac{1}{2} ||\mu_2^*||_1$$

Plugging this back into the term for the generalization risk gives the desired result:

$$R(\hat{w}) - R(w^*) \leq \gamma(n) + 2 \max_{w'} |R_{\mu^*}(w') - R_\theta(w')|$$

$$\leq \gamma(n) + 2||\mu_1^* - \theta||_1 + ||\mu_2^*||_1.$$

## B  FACTOR DEFINITIONS

We supplement the factor definitions of higher-order dependencies used in this paper. The first two stem from (Ratner et al., 2016), the rest we defined ourselves for the conducted experiments, and where motivated by frequently occurring dependency patterns, as the ones in Table 1, that are not covered by (Ratner et al., 2016). Whenever a factor $\phi_{j,k}(\lambda, y)$ is not symmetric (all factors, besides *bolstering*), we define it so that $\text{LF}_k$ acts on (e.g. *negates*) $\text{LF}_j$.

For the *fixing* dependency we have:

$$\phi_{j,k}^{Fix}(\lambda, y) = \begin{cases} +1 & \text{if } \lambda_j = -y \land \lambda_k = y \\ -1 & \text{if } \lambda_j = 0 \land \lambda_k \neq 0 \\ 0 & \text{otherwise} \end{cases}$$

for the *reinforcing* one:

$$\phi_{j,k}^{Rei}(\lambda, y) = \begin{cases} +1 & \text{if } \lambda_j = \lambda_k = y \\ -1 & \text{if } \lambda_j = 0 \land \lambda_k \neq 0 \\ 0 & \text{otherwise} \end{cases}$$

for the *priority* factor:

$$\phi_{j,k}^{Pri}(\lambda, y) = \begin{cases} +1 & \text{if } \lambda_j = -y \land \lambda_k = y \\ -1 & \text{if } \lambda_j = y \land \lambda_k = -y \\ 0 & \text{otherwise} \end{cases}$$

for the *bolstering*:

$$\phi_{j,k}^{Bol}(\lambda, y) = \begin{cases} +1 & \text{if } \lambda_j = \lambda_k = y \\ -1 & \text{if } \lambda_j = \lambda_k \neq y \lor \lambda_j = -\lambda_k \neq 0 \\ 0 & \text{otherwise} \end{cases}$$

and, finally, for the *negated* factor:

$$\phi_{j,k}^{Neg}(\lambda, y) = \begin{cases} +1 & \text{if } \lambda_j = -y \land \lambda_k = y \\ -1 & \text{if } (\lambda_j = y \land \lambda_k = -y) \lor \lambda_j = \lambda_k \neq 0 \\ 0 & \text{otherwise} \end{cases}$$

# WEAKLY-SUPERVISED GROUP DISENTANGLEMENT USING TOTAL CORRELATION

**Linh Tran**[1,2,*], **Saeid Asgari Taghanaki**[1], **Amir Hosein Khasahmadi**[1], **Aditya Sanghi**[1]
[1]*Autodesk AI Lab*
[2]*Imperial College London*

## ABSTRACT

Learning disentangled representations that uncover factors of variation in data remains an ongoing key challenge in representation learning. Recent concerns about the feasibility of learning disentangled representations in an unsupervised fashion have motivated a shift toward weak supervision. One way to incorporate weak supervision is through *match pairing*, i.e., using observations as pairs that share at least one factor of variation. Existing match pairing approaches only consider the structural constraints with an average approximate posterior over observations of a shared group. We show the limitations of these approaches and propose a novel formulation to enforce disentangled representations of groups through total correlation, which improves overall disentanglement on various image datasets.

## 1 INTRODUCTION

Decomposing data into disjoint independent factors of variations and thus learning disentangled representations is essential for interpretable and controllable machine learning Shu et al. (2019). Recent works have shown the usefulness of disentangled representation with respect to abstract reasoning (van Steenkiste et al. (2019)), fairness (Locatello et al. (2019a); Creager et al. (2019)), reinforcement learning (Higgins et al. (2017b)) and general predictive performance (Locatello et al. (2019b)). Even though unsupervised disentanglement methods (Higgins et al. (2017a); Kim & Mnih (2018); Chen et al. (2018)) have shown promising results to learn disentangled representations, Locatello et al. (2019b) showed in a rigorous study that it is impossible to disentangle variations of data without any supervision or inductive bias. Since then, there has been a shift toward weakly supervised disentanglement learning Locatello et al. (2019b), Shu et al. (2019) such as *match pairing* (Shu et al. (2019)) which uses paired observations during optimization. In this work, we present a framework to learn group-disentangled representations using total correlation in a weakly-supervised setting. Our work can be considered learning different levels of weakly-supervised group disentanglement with total correlation. Closely related work is the one of Creager et al. (2019) which proposed to minimize the mutual information between the sensitive latent variable and sensitive labels. Similarly, Klys et al. (2018) proposed to minimize mutual information between the latent variable and a conditional subspace. Both works require either supervised labels or conditions to estimate the mutual information, whereas we only use *weak* supervision for learning disentangled group representations. Locatello et al. (2020) proposed to disentangle groups of variations with only knowing the number of common groups which can be considered as a complementary component to our method. We show that our approach can flexibly disentangle between and within groups of factors of variation. Further, we demonstrate that we improve on disentanglement for various image datasets.

In summary, we make the following contributions:

1. We show limitations of existing group disentanglement approaches (Bouchacourt et al. (2018) and Hosoya (2019)) in terms of latent variable collapse and batch size sensitivity and propose a weakly-supervised way for addressing these weaknesses.

2. We propose a new way of learning disentangled representations from paired observations using total correlation. We also show how to enforce different levels of inter-group and intra-group disentanglement through total correlation.

---

[*]Corresponding author: linh.tran@autodesk.com

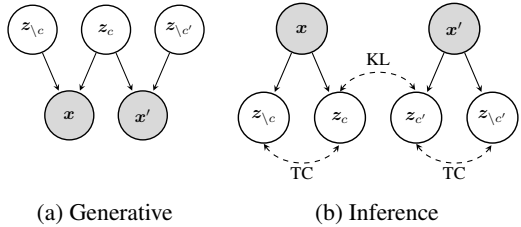

(a) Generative  (b) Inference

Figure 1: **The proposed generative and inference model.** Shaded nodes denote observed quantities, and unshaded nodes represent unobserved (latent) variables. Dotted arrows represent either minimizing the TC or the KL divergence between variables.

## 2 BACKGROUND

VAEs are latent variable models and aim to learn latent variables $z$ which should capture information about the observations $\{x_1, \ldots, x_n\}$. They are trained to maximize the evidence lower bound (ELBO) given as $\log p(x) \geq \mathbb{E}_{q_\phi}[\log p_\theta(x|z)] - D_{KL}(q_\phi(z|x) \parallel p(z))$. To be consistent with the works of Bouchacourt et al. (2018) and Hosoya (2019), let us assume that the observations $\mathcal{X} = \{x_1, \ldots, x_n\}$ are collected in $\mathcal{G}$ distinct groups. Within a group, all observations share some *fixed* factors of variations. Each group $g \in \mathcal{G}$ splits $\mathcal{X}$ into disjoint partitions with arbitrary sizes. For simplicity, we define two groups $g_C$ and $g_{\setminus C}$; where $g_C$ represents information about the actual content whereas $g_{\setminus C}$ represents any variation not contained in $C$. Given a pair of observations $(x, x')$ which share group factors $c$, we define two variables $z = (z_c, z_{\setminus c})$ and $z' = (z_c, z_{\setminus c})$ to capture content $(z_c, z_{c'})$ and non-content information $(z_{\setminus c}, z_{\setminus c'})$, e.g. style or background. (Bouchacourt et al. (2018); Hosoya (2019)) learn a group-specific latent variable $\bar{z}_{c,c'}$ by averaging over the corresponding content latent variable $z_c, z_{c'}$ during inference. The modified objective considers the ELBO of paired observations $(x, x')$ i.i.d. sampled from group $g_C$

$$\mathcal{L}_{\text{WS-ELBO}}(x, x'; \theta, \phi) = \mathbb{E}_{q_\phi}[\log p_\theta(x|\bar{z}_{c,c'}, z_{\setminus c})] + \mathbb{E}_{q_\phi}[\log p_\theta(x'|\bar{z}_{c,c'}, z_{\setminus c'})] \\ - \beta D_{KL}(q_\phi(\bar{z}_{c,c'}, z_{\setminus c}|x) \parallel p(z)) - \beta D_{KL}(q_\phi(\bar{z}_{c,c'}, z_{\setminus c'}|x') \parallel p(z)), \quad (1)$$

where $\bar{z}_{c,c'}$ is sampled from either a Normal distribution over the average of learned means and covariances (Hosoya (2019)) or a product of Normal distributions (Bouchacourt et al. (2018)).

## 3 LEARNING GROUP SIMILARITIES USING TOTAL CORRELATION

Considering the setting in Section 2, we would like 1) $z_c$ to be highly correlated with group $C$ and $z_{\setminus c}$ to be highly correlated with group $\setminus C$ and 2) $z_c \approx z_{c'}$ if the paired observations share the same content $c$ or $z_{\setminus c} \approx z_{\setminus c'}$ if the paired observations share the same non-content $\setminus c$. In what follows, we describe our approach with the generative and inference model visualized in Figure 1. Although existing works showed promising results, in practice, minimizing the objective in (1) will not necessarily fulfill the first requirement, i.e., the model will learn representations $z_c$ and $z_{\setminus c}$ that are uncorrelated with each other. Therefore, along with maximizing the variational lower bound, we propose to minimize the total correlation between latent variables $z_c$ and $z_{\setminus c}$.

The total correlation of $z_c$ and $z_{\setminus c}$ is defined as

$$\text{TC}(z_c, z_{\setminus c}) = D_{KL}(q(z_c, z_{\setminus c}) \parallel \bar{q}(z_c, z_{\setminus c})) \quad (2)$$

where $\bar{q}(z_c, z_{\setminus c})$ denotes the desired factorization of the aggregated posterior $q(z_c, z_{\setminus c})$. With different factorization, we can enforce different levels of disentanglement:

1. Inter-group disentanglement: $[z_c, z_{\setminus c}]$ is said to be disentangled if its aggregate posterior factorizes as $\bar{q}(z_c, z_{\setminus c}) = q(z_c) \cdot q(z_{\setminus c})$. Note that under this disentanglement criteria, each $z_c$ and $z_{\setminus c}$ can be correlated among themselves. However, they must be independent of each other.

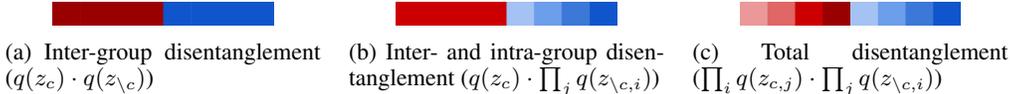

(a) Inter-group disentanglement $(q(z_c) \cdot q(z_{\setminus c}))$

(b) Inter- and intra-group disentanglement $(q(z_c) \cdot \prod_j q(z_{\setminus c, i}))$

(c) Total disentanglement $(\prod_i q(z_{c,j}) \cdot \prod_j q(z_{\setminus c, i}))$

Figure 2: **Different factorizations encourage different levels of disentanglement.** The latent variable $z_c$ representing content information is visualized as red tiles whereas $z_{\setminus c}$ representing non-content information is visualized as blue tiles. The different shades of colors represents correlation to different factors of variation.

2. Inter-group disentanglement and intra-group disentanglement of one group: $[z_c, z_{\setminus c}]$ and $z_{\setminus c}$ are disentangled if the aggregate posterior factorizes as $q(z_c, z_{\setminus c}) = q(z_c) \cdot \prod_j q(z_{\setminus c, i})$. With this criteria, $z_c$ is still free to co-vary together, but must be independent from all $z_{\setminus c, i}$. Further each dimension of $z_{\setminus c}$ is disentangled. This kind of disentanglement was also used by Creager et al. (2019).

3. Inter-group disentanglement and intra-group disentanglement of all groups: We can enforce total disentanglement if the aggregate posterior factorizes as $q(z_c, z_{\setminus c}) = \prod_i q(z_{c,j}) \cdot \prod_j q(z_{s,i})$. This is equivalent to disentanglement achieved in the FactorVAE objective Kim & Mnih (2018).

A visualization of these levels can be found in Figure 2. These types of disentanglement are useful for cases where only high-level labels are available, e.g., content vs. style, and only some groups can be further disentangled. Existing works have addressed the second requirement for group disentanglement ("shared observations have approx. same corresponding group latent variable") by using some average over the content latent variable. However, this estimate is biased and requires a certain amount of observations that share the same factors. This might be difficult with sparse, incomplete, and small datasets. We propose a KL-based regularization between the group latent variables of paired observations. This has an analytical form when the latent variables are Normal distributed and does not have any batch-size dependency. Using the total correlation (TC) and a KL term on the group latent variable, the objective becomes

$$\mathcal{L} = \sum_{\tilde{x}=x,x'} \left( \underbrace{\mathbb{E}_{q_\phi}[\log p_\theta(\tilde{x}|\tilde{z}_c, \tilde{z}_{\setminus c})]}_{\text{reconstruction}} - \underbrace{D_{\text{KL}}(q_\phi(\tilde{z}_c, \tilde{z}_{\setminus c}|\tilde{x}) \parallel p(z))}_{\text{KL between approx. posterior and prior}} \right) \\ - \sum_{\tilde{z}=z,z'} \underbrace{\beta \cdot \text{TC}(\tilde{z}_c, \tilde{z}_{\setminus c})}_{\text{total correlation}} - \underbrace{\gamma \cdot D_{\text{KL}}(q(z_g) \parallel q(z_{g'}))}_{\text{KL between shared group latent variables}}, \quad (3)$$

where $g \in \{c, \setminus c\}$ is the group which is being shared by the paired observations $(x, x')$. For evaluation, we used a binary adversary which approximates the log density ratio (Kim & Mnih (2018)) to estimate the total correlation loss. We use an adversarial network which attempts to classify between "true" samples from the aggregate posterior $q(z_c, z_{\setminus c})$ and "fake" samples from the product of the marginals $\bar{q}(z_c, z_{\setminus c})$. The latent variables are independent from each other if the samples are indistinguishable and the adversary cannot do it better than random chance.

## 4 EVALUATION

Following the experimental setup in (Locatello et al. (2019b); Chen et al. (2018)), we treated learning disentangled representations as a statistical problem instead of empirical risk minimization and hence, did not use the separate train and test sets. For evaluation, we used two datasets, namely, 3DShapes (Burgess & Kim (2018)) and dSprites (Matthey et al. (2017)). We compare our model, *group-tcVAE*, with MLVAE, Bouchacourt et al. (2018), and GVAE, Hosoya (2019). Locatello et al. (2020) have already shown that both works by Bouchacourt et al. (2018) and Hosoya (2019) are superior to the unsupervised disentanglement approaches, hence, we do not compare with them. We quantitatively compare the strength of disentanglement with the Mutual Information Gap (MIG) (Chen et al. (2018)). Further, we introduce *group-MIG*, a metric based on MIG, which quantitatively estimates the mutual information between groups and corresponding latent variables. Formally, we define group-MIG as $\frac{1}{K} \sum_{k=1}^{K} \frac{1}{H(v_k)} (\max I(z_{i=f_g(v_k)}; v_k) - \max I(z_{i \neq f_g(v_k)}; v_k))$ where $K$ is the

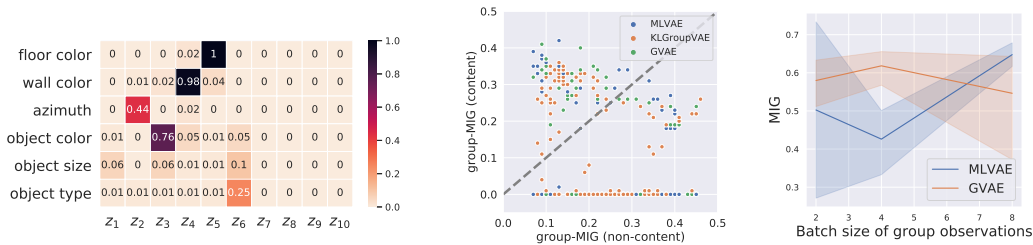

(a) 3DShapes: MI between latent dimensions and factors of variation of trained GVAE model with MIG = 0.55 and group-MIG = 0.44.

(b) dSprites: group-MIG of content and non-content information for all hyperparameter runs for MLVAE and GVAE.

(c) dSprites: MIG w.r.t. different number of shared observations for MLVAE and GVAE.

Figure 3: **Collapse and sensitity of existing weakly-supervised disentanglement models.** In all the sub-figures higher is better.

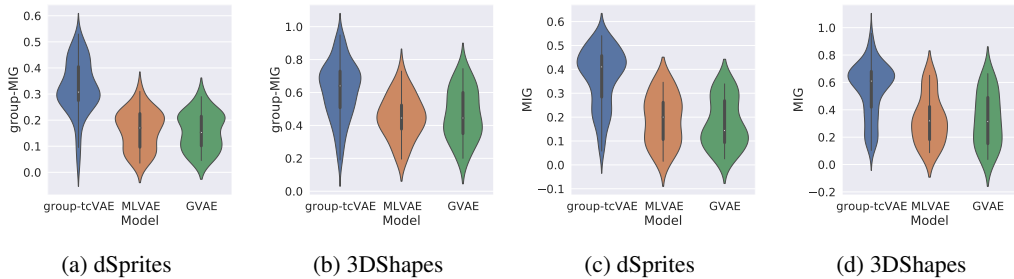

(a) dSprites  (b) 3DShapes  (c) dSprites  (d) 3DShapes

Figure 4: **Comparisons between group-tcVAE and comparisons.** Density plots of group-MIG and MIG for group-tcVAE, MLVAE and GVAE) over all runs (*higher is better*).

number of known factors, $v_k$ is the ground truth factor, $f_g(v_k) \in \{c, \backslash c\}$ returns the group that the factor belongs to and $I(z; v_k)$ is an empirical estimate of mutual information between continuous variable $z$ and $v_k$.

### 4.1 EMPIRICAL ANALYSIS OF EXISTING WEAKLY-SUPERVISED METHODS

**Collapse of content latent variable.** The latent variable $z_c$ can either collapse to a single factor of variation or it might even contain almost no information to any content. We visualize such behavior in Figure 3 (a) on a GVAE model trained on 3DShapes with two groups of variations $c =$ {object color, object size and object type} and $\backslash c =$ {floor color, wall color, azimuth}. Ideally, $z_1 - z_5$ contains high mutual information with group factors $\backslash c$ and $z_6 - z_{10}$ contains high mutual information with group factors $c$. However, most information is captured in $z_1 - z_5$, whereas only a little information about object type is contained in $z_6$. As shown in Figure 3 (b), we make similar observations with dataset dSprites in which both MLVAE and GVAE fail to capture content-specific information in the corresponding latent variable.

**Sensitivity to group batch size.** In practice, always having a certain number of observations that share the same group variations might be difficult, which is a requirement for MLVAE and GVAE with dSprites. This results in performance degradation and high variance (Figure 3(c)).

### 4.2 WEAKLY-SUPERVISED DISENTANGLEMENT

We perform an extensive evaluation on group-tcVAE to assess its performance in comparison to MLVAE and GVAE. For MLVAE and GVAE, we experimented with the hyperparameters as in Locatello et al. (2020). For group-tcVAE, we used the hyperparameter ranges $\beta = [10, 20, 30, 40, 50, 100]$, $\lambda = [1, 8, 16, 32, 64]$ and a batch size of 32 paired observations ($= 32 \times 2$). For all models, we performed five runs with different random seeds. We plotted all results in Figure 4. For both group-MIG

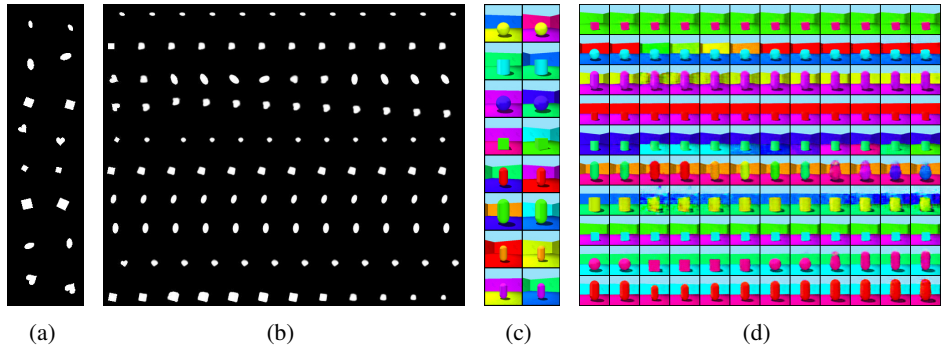

Figure 5: **Qualitative results of group-tcVAE.** Training samples (a, c) and reconstruction as well as interpolations (b, d) for dSprites (a, b) and 3DShapes (c, d). Each row represents a pair of observations in (a, c). For the interpolations, each i-th row represents interpolating over only the i-th dimension of $z$.

and MIG, group-tcVAE outperforms MLVAE and GVAE w.r.t. average and best MIG and group-MIG. With dSprites, group-tcVAE almost doubles the average and best performance, whereas with 3DShapes we observe an increase of at least 10% w.r.t. group-MIG and MIG. For the best performing models, we also plotted training samples and qualitative results in Figure 5.

## 5  CONCLUSION

We have analyzed existing weakly-supervised disentanglement models and identified challenges w.r.t. latent variable collapse and batch size sensitivity. We proposed a new framework based on total correlation for weakly-supervised disentanglement and showed through empirical evaluations on image datasets that our model improves learning disentangled representations. For future work, we plan to apply our proposed framework to challenging real-world data sets and non-image domains and extend it to semi weakly-supervised and active learning settings.

# USING SYSTEM CONTEXT INFORMATION TO COMPLEMENT WEAKLY LABELED DATA

**Matthias Meyer & Lothar Thiele**
Computer Engineering and Networks Laboratory
ETH Zurich
Zurich, Switzerland
{matthmey,thiele}@ethz.ch

**Michaela Wenner & Fabian Walter**
Laboratory of Hydraulics, Hydrology and Glaciology
ETH Zurich/WSL Birmensdorf
{wenner,walter}@vaw.baug.ethz.ch

**Clément Hibert**
Institut de Physique du Globe de Strasbourg
University of Strasbourg
hibert@unistra.fr

## ABSTRACT

Real-world datasets collected with sensor networks often contain incomplete and uncertain labels as well as artefacts arising from the system environment. Complete and reliable labeling is often infeasible for large-scale and long-term sensor network deployments due to the labor and time overhead, limited availability of experts and missing ground truth. In addition, if the machine learning method used for analysis is sensitive to certain features of a deployment, labeling and learning needs to be repeated for every new deployment. To address these challenges, we propose to make use of system context information formalized in an information graph and embed it in the learning process via contrastive learning. Based on real-world data we show that this approach leads to an increased accuracy in case of weakly labeled data and leads to an increased robustness and transferability of the classifier to new sensor locations.

## 1 INTRODUCTION

Classifiers based on artificial neural networks have proven to be very effective across domains, however their applicability to real-world data is limited by the requirement of a clean and comprehensive dataset (Tsipras et al., 2020). Unfortunately, real-world datasets often contain artefacts arising from the system environment and contain incomplete and uncertain labels. One example of machine learning applications is natural hazard monitoring for slope failure detection (Hammer et al., 2013; Dammeier et al., 2016). Here, high misclassification requires careful retraining and post-processing (Hibert et al., 2017). In this setting, comprehensive manual annotations are infeasible for large-scale and long-term sensor network deployments due to the labor and time overhead (Meyer et al., 2019).

Hence, the process is error-prone and requires significant domain expertise. However, experts might not be available throughout the whole deployment periods of the sensor network, which inevitably leads to an annotation set containing noisy annotations limited in time and/or subset of sensors. In addition, as long as the learned features and classifier are sensitive to the detailed properties of the subsurface and the sensors, labeling and learning needs to be repeated for every new installation or classifier performance is decreased (Wenner et al., 2021). Therefore, there is a close link between weakly labeled data and robustness with respect to certain feature variations.

Fortunately, real-world deployments provide additional sources of information which could be beneficial for learning, such as correlation of sensor data due to sensor proximity. However, these information cannot be easily captured by the prevailing data/annotation pairs used for learning. Similarity learning (Schroff et al., 2015; Meyer et al., 2017), such as contrastive learning (He et al., 2020; Chen et al., 2020; Saeed et al., 2020) allows to establish relations between data pairs. However, their capability to integrate system context information is limited.

To address these challenges, we propose to transfer the concept of knowledge graphs (Hogan et al., 2021) to learning by using it for storing information about data similarity. Moreover, we extend the prevailing data/annotation learning concept to allow any data point to be an annotation for any other data point. This is accomplished by utilizing the following concepts: (i) injecting all available knowledge in form of an information graph and sampling from it, (ii) transforming the data into a common representation and (iii) the use of contrastive learning to train the system. We show that using these concepts to formalize system context information and using the additional knowledge in the learning phase leads to an increased accuracy in case of weakly labeled data and leads to an increased robustness and transferability of the classifier to new sensor locations. [1]

Our main contributions are:

- We present a method which uses system context information to counteract the negative impact of few and weak labels by combining contrastive learning with an information graph.
- We present a unified learning process in which annotations are encoded as Gaussian random vectors to treat them similar to data.
- We demonstrate on a dataset gathered from a real-world deployment in the Swiss alps, how the method can be used to train a classifier with improved generalization performance across sensors with diverging characteristics.

## 2 DATASET

In this work, we use data from a real-world deployment of seismic sensors at Illgraben, Switzerland. The sensor array consists of 8 seismometer (ILL01-08), each having three channels, one vertical and two horizontal. The sensors are deployed at distances of hundreds of meters up to several kilometers away from the area of interest. We aim to distinguish seismic signals from 3 different types of events namely earthquakes, slope failures and noise signals. The Illgraben event catalog was created by visual inspection of the vertical channel of the seismic waveforms and their spectrograms by experts. The earthquake catalogs provided by the Swiss Seismological Service (SED) and the European-Mediterranean Seismological Center (EMSC) served as additional ground truth for providing correct earthquake labels. The Illgraben event catalog consists of 320 to 560 time segments per station each containing an event, summing up to 32.5 hours of labelled seismic data recorded at a sampling frequency of 100 Hz. In addition, the dataset contains randomly sampled, verified time segments without activity with a total duration approximately equal to the event segment's total duration.

## 3 METHOD

In our scenario, two major issues need to be addressed, namely (i) few and weak labels and (ii) classifier robustness. The first issue requires an improvement and extension of the annotation set, the latter requires that the learning method can adapt to out-of-distribution samples.

In contrast to real-time classification, environmental monitoring usually relies on post-processing of a long-term dataset. Therefore, an extensive dataset is usually available for training albeit not always thoroughly annotated. In our scenario, we can make use of non-annotated data by using general assumptions about the specific sensor deployment, for example about sensor proximity: The same event is captured by multiple seismometer channels and possibly multiple stations, but with different signal signatures. These differences are caused by groundwave propagation as well as properties of the seismometers, for example ground coupling. Thus, we obtain "different views" of the same event. Contrastive learning has shown to benefit from such different views. Intuitively speaking, contrastive learning achieves an embedding of data samples in a latent space by moving representations of different views of the same event closer together while increasing the distances of representations of different events.

To combine contrastive learning with all available information, we propose to make use of an information graph, which holds annotations as well information about the relation between time segments, channels and stations. We expect that by using system context information we simultaneously

---

[1] Further content is available at https://matthiasmeyer.xyz/system-context-info/

(i) improve our annotation set by adding additional information to each annotated segment and (ii) include non-annotated, out-of-distribution samples in the training process.

The information graph is filled by subdividing the seismic signals into segments using a window length $T_w$. Each segment is represented by a node in the information graph. An edge is introduced between segments A and B if the segments overlap in time and A is from a different station or different channel than segment B. To reduce the possibility of learned shortcuts (Fonseca et al., 2020) no edges to segments of the same channel are added.

The information graph is used to train a model $f(\cdot)$ which embeds each segment $x_i$ into a common space $z_i = f(x_i)$, with $z_i \in \mathbb{R}^d$. Similar to related work, we separate the model into an encoder and encoder head (Chen et al., 2020). As illustrated in Fig. 1, a fixed number $N_e$ of edges is sampled from the graph for every batch during training and connected data segments are loaded. Any duplicate segments are removed from the batch before computing $f(\cdot)$, leading to a number of data segments in the batch of $N \leq 2N_e$. Each data segment is encoded by the encoder and subsequently transformed by the encoder head into an embedding vector $z_i$.

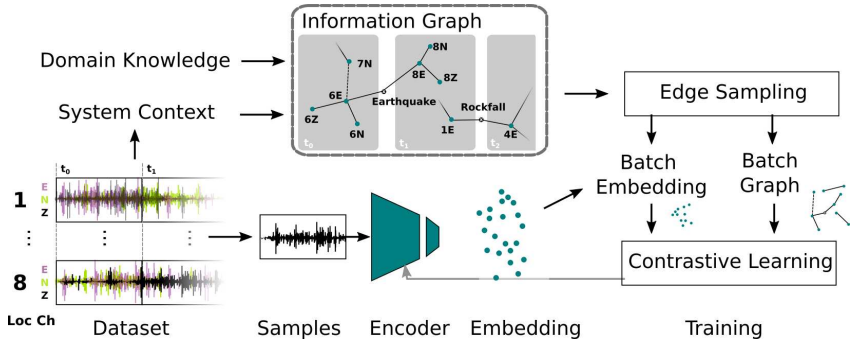

Figure 1: The central entity is the information graph which combines knowledge from domain experts, for example annotations or signal propagation behaviour, as well as dataset-specific knowledge of each data segment, for example **loc**ation, **ch**annel, time. The combination of contrastive loss, information graph and encoder allows to learn a suitable embedding for the classification task.

By sampling the edges we construct a subgraph of the information graph with non-negative adjacency matrix $A \in \mathbb{R}_{\geq 0}^{N \times N}$. To avoid that second order neighbours of a node have a detrimental impact on training by not being directly connected, we add second order neighbors by $B = A + A^2$. In this setting, we define the contrastive loss between a pair $s, t$ (source and target of an edge), with the adjacency matrix $B$ as follows:

$$L_{s,t} = -B_{s,t} \log \frac{\exp(\phi(z_s, z_t)/\tau)}{\sum_{n=1}^{N} \mathbb{1}_{[B_{s,n}=0]} \exp(\phi(z_s, z_n)/\tau)} \quad (1)$$

where $B_{s,t}$ represents the weight of the edge connecting $s, t$. $\phi(\cdot)$ is a similarity function, which in our implementation is the Cosine similarity. $\tau$ is a temperature scaling. The indicator function $\mathbb{1}$ is evaluating to 1 iff $B_{s,n}$ is zero.

Annotations are considered in the information graph by introducing $N_c$ anchor nodes, where $N_c$ equals the number of classes. Each segment belonging to a class is connected to the anchor node of that class by an edge. Two strategies can be employed to compute Eq. 1 for an edge with an annotation anchor node.

The first option makes use of the fact that $B$ contains second order neighbors meaning that all nodes sharing an edge with an annotation are also directly connected. Thus, the edges with an annotation node can be skipped while computing Eq. 1 but representations with the same annotation are still moved closer together, which resembles the work by Khosla et al. (2020).

The second option is to introduce a high-dimensional L2-normalized Gaussian random vector $a^{(c)} \in \mathbb{R}^d$ for class $c$ into the batch which acts as the target $z_t$ during computation of Eq. 1. The vectors

|  | Accuracy | | | |
|---|---|---|---|---|
|  | all | all+SC | one-station | one-station+SC |
| Random Forest | 86.4 % | - | n.a. | - |
| ResNet18+XE | 91.3 % ±0.5 | - | 50.0 % ±15.0 | - |
| ResNet18+IG+links | 92.3 % ±0.3 | 93.8 ±0.3 | 44.0 % ±15.6 | 84.1 % ±2.6 |
| ResNet18+IG+anchors | 92.0 % ±0.4 | 93.9 ±0.6 | 47.9 % ±16.7 | 85.0 % ±1.8 |

Table 1: Classifier accuracies for different sets of available training annotations. Either all annotations (*all*) or only annotations for one station are available (*one-station*). Additionally, the information graph (IG) is used with or without system context information (*SC*).

are fixed at the beginning of the training. Here, we make use of the fact that any two random high-dimensional vectors are almost orthogonal to each other with high probability (Blum et al., 2020), thus data points of different classes are trained to move "far away" from each other. The first strategy will be referred to as *link*, the latter as *anchor* in the following evaluation.

## 4 EXPERIMENTAL EVALUATION

We evaluate the proposed approach by performing an ablation study and comparing the system to a random forest classifier, which is best practice for slope failure detection (Wenner et al., 2021).

For the ablation study we use a classifier based on a single-channel variant of ResNet18 (He et al., 2015) as encoder in combination with an MLP with 1 hidden layer as encoder head. During classification the encoder head is replaced with a classification head differing only in output size. Input to the ResNet18 is a log-compressed spectrogram of the seismic data. For more implementation details please refer to the Appendix A. The ResNet18 model is trained with three methods, cross-entropy loss between the output of the classification head and the ground truth (Resnet18+XE), contrastive pretraining using the information graph (IG) and either the link (Resnet18+IG+link) or anchor (Resnet18+IG+anchor) strategy. Subsequently, the classification head is trained using cross-entropy loss. We compare training with system context information (SC) and training without it.

The benefit of our approach for the weakly-labeled setting is evaluated by using the available training annotations of all stations. The experiments are repeated 5 times and mean and standard deviation are reported. Robustness is evaluated by training the model variants when only a subset of the annotations are available. While the whole training data is available to train a classifier only the annotations for one of each of the 8 seismic stations can be used. All reported accuracies are based on evaluation on the test set using all stations and are reported as mean and standard deviation of all one-station evaluations.

The results presented in Table 1 show that the ResNet18 classifiers outperform the random forest classifier in the weakly-labeled setting (*all* and *all+SC*). The *all* column, illustrates that training using contrastive pretraining improves the performance significantly in comparison to the random forest classifier but only slightly in comparison to using cross-entropy loss (ResNet18+XE). However, if we include the system context information the accuracy improves significantly (*all+SC* column), demonstrating our method's applicability to weakly labeled data.

In the robustness experiments, all classifiers show a comparable bad performance of around 50 % on average if only annotated data of one station is used (*one-station*). If more non-annotated data from other stations is available, our method takes advantage of the system context information (SC) stored in the information graph (*one-station+SC*) and the average accuracy rises to over 84 %. The increase comes from a better generalization to other sensors, as illustrated in Fig. 2. The left figure illustrates the poor generalization of ResNet18+XE to other stations than the one used for training. If, however, non-annotated data from other stations and system context information is available, our method increases classifier performance on all stations, thus demonstrating and increased robustness.

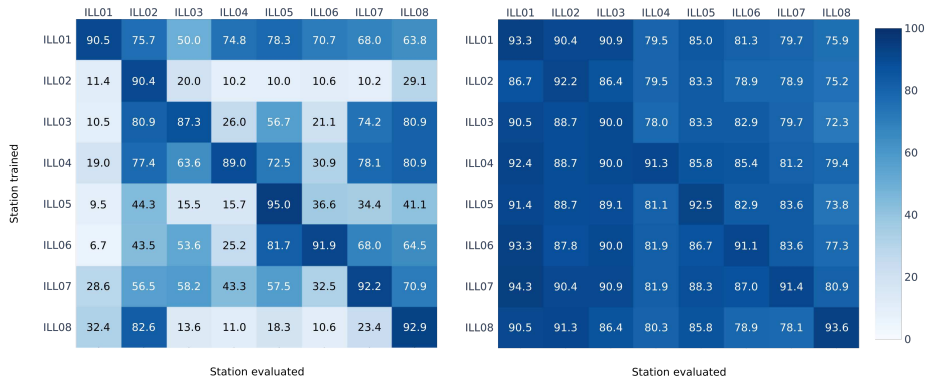

Figure 2: Each row indicates of which station the annotated training subset was used. Each column indicates the respective score on the subset of the test dataset for each station. (Left): Results for ResNet18+XE. (Right): Results for ResNet18+IG+anchor.

## 5 CONCLUSION

In this paper we have presented a novel approach to learn with weakly labeled data for the case of mass movement monitoring. By using contrastive learning we can increase the classification accuracy compared to the reference implementation. Moreover, the presented method unifies data and annotation representations and thus inherently allows to integrate additional system information into the learning process. This additional information leads to a strong performance increase in a setting with limited annotations and diverging sensor characteristics, demonstrating increased robustness across sensors.

Published as a workshop paper at the Workshop on Weakly Supervised Learning / ICLR 20216

Table 2: Random forest parameters

| | |
|---|---|
| Number of trees | 400 |
| Split quality measure | Gini criterion |
| Minimum number of samples required to be a leaf node | 1 |
| Minimum number of samples for an internal node to be split | 2 |

## A  IMPLEMENTATION DETAILS

### A.1  DETAILS TO CONTRASTIVE LEARNING WITH INFORMATION GRAPH

The seismic signals are subdivided into segments using a window length $T_w$ and a stride $T_h$. The data subset for pre-training uses $T_w = 30s$, $T_h = 30s$, the subset for fine-tuning and the test set use $T_w = 30s$, $T_h = 15s$. Each segment's annotation is determined using the Illgraben event catalogue, which is split into training and test set with a ratio of approx. 70/30. Linear detrend is applied to the seismic signal before it is transformed into a log-compressed spectrogram with window length of 2.56 s and stride of 0.08 s. No data augmentation is applied. As encoder we use a single-channel variant of ResNet18 (He et al., 2015), without the final linear layer. The output of the encoder is a 512-dimensional vector, which is then passed through the encoder head, consisting of a MLP with a hidden layer of size 512, batch normalization and ReLU non-linearity. The encoder head's output size is $d = 128$ and L2 normalized. During fine-tuning, the same encoder head architecture (with random initialization) is used as a classification head with an output size equal to the number of classes $N_c$.

In the supervised training we train encoder and classification head jointly using cross-entropy loss.

For semi-supervised training (*all+SC* and *one-station+SC*), we train the encoder and encoder head with contrastive loss, then for fine-tuning we replace the encoder head with a classification head and train the classification head with cross-entropy loss while keeping the encoder weights fixed. In every epoch we first train with contrastive loss, then fine-tune the classification head. For optimization we use SGD with momentum 0.9 and weight decay $10^{-4}$ and a batch size of 128. The temperature coefficient $\tau$ is set to 0.1. We use a cosine annealing scheduler. In our experiments the edge weights of the information graph are 1.

To counter class-imbalance, each batch contains the same number of examples for each class. During semi-supervised training the non-annotated data out-weights the annotated data by a factor of approx. 4.5 in each batch. We select our model based on a validation set which is 20% of the training set, except for the one-station+SC experiment. Here, we select model from the last epoch, since model selection on the one-station subset would deteriorate the generalization effect. More details, e.g. the code and hyperparameters for individual experiments will be made available on the paper's project page.

### A.2  DETAILS TO RANDOM FOREST CLASSIFIER

FollowingProvost et al. (2017) and Wenner et al. (2021) we computed a total of 55 signal characteristics in the time and frequency domain, e.g., information on the signal form and dominant frequencies. For a complete description of the chosen features see Wenner et al. (2021). We performed a three-fold-cross-validation grid search to optimize classifier performance. Final parameters are presented in Table 2.



\end{document}